\documentclass[10pt,journal,compsoc]{IEEEtran}
\ifCLASSOPTIONcompsoc
  \usepackage[nocompress]{cite}
\else
  \usepackage{cite}
\fi

\ifCLASSINFOpdf
  \usepackage[pdftex]{graphicx}
\else
  \usepackage[dvips]{graphicx}
\fi

\usepackage{xcolor}
\usepackage{multirow}
\usepackage{amsmath,amssymb}

\DeclareMathOperator*{\round}{round}
\usepackage{array}

\ifCLASSOPTIONcompsoc
  \usepackage[caption=false,font=footnotesize,labelfont=sf,textfont=sf]{subfig}
\else
  \usepackage[caption=false,font=footnotesize]{subfig}
\fi

\begin{document}

\title{KitBit: A New AI Model for Solving Intelligence Tests and Numerical Series}
\author{Víctor~Corsino, %\orcidlink{0000-0003-3456-9871},
        José~Manuel~Gilpérez, %\orcidlink{0000-0001-5612-0122},
        and~Luis~Herrera
        
\IEEEcompsocitemizethanks{\IEEEcompsocthanksitem V. Corsino and J.M. Gilpérez are with School of Industrial and Aerospace Engineering, Universidad de Castilla La-Mancha, Avenida Carlos III s/n, Real Fábrica de Armas, 45071 Toledo, Spain.
\IEEEcompsocthanksitem L.A. Herrera is CEO at \textregistered{Smart Technologies Investments}.}
\thanks{Manuscript received November 02, 2022; revised November 22, 2022.}}
% The paper headers
\markboth{IEEE TRANSACTIONS ON PATTERN ANALYSIS AND MACHINE INTELLIGENCE,~Vol.~45, No.~11, November~2023}%
{Corsino\MakeLowercase{\textit{et al.}}: KitBit: An AI Cognitive Model for Solving Intelligence Tests and Numerical Series}
% The only time the second header will appear is for the odd numbered pages after the title page.

\IEEEtitleabstractindextext{%
\begin{abstract}
The resolution of intelligence tests, in particular numerical sequences, has been of great interest in the evaluation of AI systems. We present a new computational model called KitBit that uses a reduced set of algorithms and their combinations to build a predictive model that finds the underlying pattern in numerical sequences, such as those included in IQ tests and others of much greater complexity. We present the fundamentals of the model and its application in different cases. First, the system is tested on a set of number series used in IQ tests collected from various sources. Next, our model is successfully applied on the series used to evaluate the models reported in the literature. In both cases, the system is capable of solving these types of problems in less than a second using standard computing power. Finally, KitBit's algorithms have been applied for the first time to the complete set of entire series of the well-known OEIS database. We find a pattern in the form of a list of algorithms and predict the following terms in the largest number of series to date. These results demonstrate the potential of KitBit to solve complex problems that could be represented numerically.
\end{abstract}

}

\maketitle

\IEEEdisplaynontitleabstractindextext
\IEEEpeerreviewmaketitle

%%%%%%%%%%%%%%%%%%%%%%%%%%%%%%%%%%%%%%%%%%%%%%%%%%%%%%%%%%%%%%%%%%%%%%%%%%%%%%%%%%%%%%%%%%%%%%%%%%%%%%%%%%%%%%%%%%%%%%%%%%%%%%%%%%%%%%%%

\IEEEraisesectionheading{\section{Introduction}\label{sec:introduction}}

\IEEEPARstart{P}{attern} recognition is a main aspect of any intelligent system, and is therefore of great interest to AI \cite{LIU}. Pattern recognition is defined as the search for regularities and structure in data \cite{BEZ}, and is a fundamental part of machine learning. Computational models oriented to pattern recognition have been applied in many areas, such as image processing or voice recognition, and is currently applied with great success thanks to techniques such as Deep Learning. In particular, pattern recognition has traditionally been used to assess inductive reasoning skills on IQ tests, and theories explaining the strategies employed by humans to solve them have been proposed \cite{SIMKOT}, \cite{KOTSIM}, \cite{HOLPELLGLA}. It is natural that this type of problem is of interest in AI, where different models have been developed to solve IQ tests \cite{HWD}, \cite{SR}, \cite{HMSSD} and the prediction of numerical sequences has been proposed as a method to evaluate the computational capabilities of machine learning models \cite{NAM}.

This work presents a new model for pattern recognition called KitBit, which successfully solves the largest number of numerical series reported to date through a computer system. Our approach is close to the proposals of Simon and Kotovsky \cite{SIMKOT}, \cite{KOTSIM}, and Holzman, Pellegrino and Glaser \cite{HOLPELLGLA}, who developed a theory to explain how humans carry out the task of infer, remember, and extrapolate patterns in sequences of letters or numbers. Their investigations established a general model consisting of a four-stage process: i) relationship detection, ii) periodicity discovery, iii) pattern description and iv) extrapolation. The first two involve discovering the pattern or structure inherent in the series. The third consists of remembering the description of the pattern through a certain language. And the fourth supposes to reproduce and extrapolate the sequence from the previous description. Our model performs the previous stages, using an improved strategy in the first stage of relationship detection. According to LeFevre and Bisanz \cite{LEFBIS}, the participants in the trials explore the series to postulate and test hypotheses about the relationships between elements that occupy adjacent positions \cite{LEISTE} fundamentally using two procedures, the activation of the semantic memory and calculation. In complex series, the difficulty depends on the arithmetic operations involved \cite{SIMKOT}. Likewise, the previous knowledge of the participants is also a relevant factor \cite{LEFBIS}, having shown that learning a certain procedure allows for significantly improving the results \cite{SCHSPA}. In this sense, KitBit incorporates, as part of the first stage, the so-called difference tables, which have been suggested as a simple and efficient way to solve numerical IQ tests \cite{DEW}. The rest of the operations performed by the human to transform the series in this stage, such as symmetry, analogy or focusing, are also performed in our model \cite{DIEMIC}. This can be illustrated in the following stages: the human describes the pattern through words and symbols or through a recurrence formula, while the program establishes a pattern description language based on sequences of procedures. Sequence extrapolation is also performed respectively in both cases, obtaining new terms by applying the already known pattern. The result identifies emergent concepts with new properties not found in the sum of the lower-level set of concepts.

Therefore, the differences between the procedure carried out by the humans and KitBit fundamentally lie in the resources put into play to detect the relationships between the terms of the series. On the one hand, the human is limited by short-term memory and the type of operations that can be performed mentally \cite{HOLPELLGLA}, \cite{LEFBIS}. On the other hand, instead of exclusively analyzing the series, the model also evaluates the new series generated from successive variations of the original series. This is represented in the difference table using a sequence of actions or \textit{kitas}. This collection of \textit{kitas} adds other operations beyond basic arithmetic, such as powers, exponentials, logarithms, etc., and constitutes a \textit{toolbox} of procedures to find patterns, which are iteratively concatenated to be applied to the variables that define a system or their combinations. Through these significant methodologies, the system effectively discover regularities in complex numerical series, beyond those that a person could undertake during the execution of an IQ test.

Our model differs from symbolic regression where a recurrence function or relation is inferred from a sequence of numbers. To date, symbolic regression using neural networks has only been applied to simple cases with very limited results \cite{DSR}. Using KitBit, instead of a mathematical function, the underlying pattern is represented by a sequence of \textit{kitas}, which are performed on the elements of the series. In this way, KitBit analyzes the problem from different perspectives through a dynamic process of approximation and framing. This results in the underlying pattern in the form of a sequence of operations that can be easily identified and stored. Once the pattern is known, new terms can be predicted. KitBit algorithms have been used to find the implicit patterns in the integer sequences of the complete Online Encyclopedia of Integer Sequences (OEIS) database \cite{OEIS}, \cite{OEISD}, many of which are highly complex and reach the largest number of resolved series reported to date. Likewise, KitBit is capable of solving practically all of the numerical series that are present in IQ tests.

%%%%%%%%%%%%%%%%%%%%%%%%%%%%%%%%%%%%%%%%%%%%%%%%%%%%%%%%%%%%%%%%%%%%%%%%%%%%%%%%%%%%%%%%%%%%%%%%%%%%%%%%%%%%%%%%%%%%%%%%%%%%%%%%%%%%%%%%%%%%%%%%%%%%%%%%%%%%%%%%%%%%%%%%%%%%%%%%%%%%%%%%%%%%%%%%%%%%%%%%%%%%%%%%%%%%%%%%%%%%

\section{Resolution of IQ test and numerical series: previous models}

Because IQ is considered to be a predominant benchmark for measuring human intelligence \cite{CPR}, \cite{DHO}, IQ tests are an interesting challenge for AI systems. They are also a useful tool to quantify how similar these systems are to human intelligence \cite{CPC}. Among the many different types of tests, IQ tests based on numerical sequences are fundamental because the other tests can be transformed into a numerical problem. These problems can be defined mathematically by means of a function that assigns the natural numbers to the real numbers: $f:\mathbb{N} \rightarrow \mathbb{R}$ where each element is defined algorithmically. In addition, they are normally based on simple patterns or a combination of patterns, where the numbers are restricted to integer values that are generally small enough to allow easy mental calculations using the four basic arithmetic operations. The number of elements is also relatively small, usually no more than six to eight terms.

Few publications are related to the computational resolution of the IQ test in the conventional literature on AI \cite{DEEPIQ}. In 1963, Evans \cite{EVA}, and Simon and Kotovsky \cite{SIMKOT} designed AI programs that were capable of identifying regularities in patterns in analogy problems and letter series termination problems, respectively \cite{HAC}. In the 1960s and 1970s, Fredkin, Pivar and Finkelstein, and Pivar and Gord, in early research on inductive learning, developed LISP programs that operated on integers and automatically discovered interesting relationships in the data but could only work with a limited class sequence \cite{CSDM}. In 1980, Hofstadter developed a series of computational models in the Copycat project with the main objective of understanding analogy \cite{HAC}. Between 2006 and 2011, computational models aimed at solving IQ test problems started to become more popular, either trying to understand human cognition or as a method to evaluate AI techniques \cite{HAC}. Meanwhile, between 2011 and 2014, several models were proposed to solve IQ tests based on numerical series, although some of them performed (in general) worse than human beings \cite{HWD}.

More recently, Liu et at. \cite{HWD} conducted a comparison study of different state-of-the-art approaches to automatically solve an IQ test. The authors collected an extensive dataset of these tests, containing 10,000 questions of different types, where over 2,500 are numerical sequences. Meanwhile, other studies have focused more generally on numerical series datasets. For instance, Ragni and Klein \cite{RAG} tested their model, which was based on an artificial neural network (ANN), on a series that was selected from OEIS. At that time, the OEIS database contained 187,440 number series, from which they selected 57,524 that consist of at least 20 numbers with values $\pm 1000$. Using this approach, the authors solved 26,951 of the selected number series. Meanwhile, Siebers and Schmid \cite{SNNSI} avoided the use of the OEIS database to test their semi-analytical cognitive model, considering that most of the OEIS series are too complex to be induced by humans. Instead, they used a random generator \cite{AIIS} to construct 25,000 number series using addition, subtraction, division, multiplication and exponentiation. Based on this dataset, the hit rate to induce the next three numbers in the series is 93.2\%. 

Some approaches use a very limited number of series to test their models, which are summarized in the reviews provided by Schmid and Ragni \cite{SR}, and Hernández-Orallo et al. \cite{HMSSD}. The latter provides a complete account of 30 computer models that are focused on solving intelligence tests, reviewing their purpose, their degree of specialization and the techniques used, as well as their achievements and the relationship between them. Meredith \cite{SW} evaluated the Seek-Whence model using only 12 numerical series (i.e., the Blackburn dozen) to solve the Bongard problem \cite{PR}. This approach uses pattern recognition and analogy according to the ideas of Hofstadter \cite{FCCA} without requiring typical mathematical operations. Using a similar approach, Mahabal \cite{MTH} developed SeqSee (i.e., a computer program that emulates intelligent activities) to extend integer sequences, while avoiding the use of brute force or computing shortcuts that are implausible in humans. Mahabal analyzes 146 selected numerical series related using the group of intelligent activities emulated. Sanghi and Dowe \cite{CPC} presented a simple computer program that was able to solve IQ tests focused on pattern recognition of certain types of numerical series, such as arithmetic progression, geometric progression, Fibonacci series or powers of a series. Their results are based on 12 IQ test, which are not presented. Burghardt \cite{EGUG} applied an algorithm that was called E-generalization to discover the pattern underlying numerical series. The results were presented in nine series, including alternating and Fibonacci series. Stranneg{\aa}rd et al. \cite{SAU} implemented ASolver, which is a rule-based system that includes memory restrictions that are similar to human capabilities. The system's performance was evaluated with 11 (non-published) IQ tests. Hofmann et al. \cite{HKS} developed the IGOR2 analytical inductive programming system and compared its achievements with the approaches of Ragni and Klein \cite{RAG}, and Burghardt \cite{EGUG} based on 20 and eight selected series, respectively. Finally, the authors selected 100 series, some of which were included in OEIS, to test the model. Recently, Ryskina and Knight \cite{LMPI} used NPL models trained on OEIS to complete 57 integer series collected from online IQ test preparation websites.

Although the previous models have tried to induce the next numbers in the series, they do not attempt to obtain a symbolic expression from the data. Recently, D'Ascoli et al. \cite{DSR} trained a neural network to infer the symbolic regression of integer sequences. They evaluate their model on a subset of 10000 series from the OEIS database, which were labeled as `easy' \cite{WU}. For this subset, the model reaches an accuracy of 53\% at next term prediction but was only 27\% accurate when predicting the next 10 terms. For the symbolic regression, the model achieves an accuracy of 19\%.

In summary, the computational models that have been created to induce a pattern in a numerical series use only a very small number of examples or select their examples with certain restrictions to test the model. In contrast, in this paper we will show how KitBit tries to solve the complete set of entire sequences of the OEIS database, as well as all the numerical series used to evaluate the different models found in the literature, to find a pattern in the largest number of numerical series of different types reported to date. This will lay a solid foundation to address other types of problems.

%%%%%%%%%%%%%%%%%%%%%%%%%%%%%%%%%%%%%%%%%%%%%%%%%%%%%%%%%%%%%%%%%%%%%%%%%%%%%%%%%%%%%%%%%%%%%%%%%%%%%%%%%%%%%%%%%%%%%%%%%%%%%%%%%%%%%%%%%%%%%%%%%%%%%%%%%%%%%%%%%%%%%%%%%%%%%%%%%%%%%%%%%%%%%%%%%%%%%%%%%%%%%%%%%%%%%%%%%%%%

\section{The KitBit model} \label{s4}

The KitBit model is based on four fundamental components: a basic data storage unit or \textit{edk}, a \textit{kitas} or actions carried out on the \textit{edk}, a pattern search system and a new elements prediction system. KitBit's algorithms are coded using the Python programming language and they are applied to solve numerical series problems by finding their underlying pattern and allowing new terms of the series to be inferred.

%%%%%%%%%%%%%%%%%%%%%%%%%%%%%%%%%%%%%%%%%%%%%%%%%%%%%%%%%%%%%%%%%%%%%%%%%%%%%%%%%%%%%%%%%%%%%%%%%%%%%%%%%%%%%%

\subsection{The basic unit or \textit{edk}} 

For a series of numbers $X = \{x_{1},x_{2},...,x_{n}\}$, where $x \in \mathbb{R}$ and $n \in \mathbb{N} \hspace{0.1cm} | \hspace{0.1cm} n\geq 2$, the \textit{edk} is constructed as a difference table or a ratio table \cite{AIS}.

\begin{equation}
\label{eqedk}
edk_{n} = \begin{IEEEeqnarraybox*}[][c]{,c/c/c/c/c/c,}
y_{1}^{n-1} & & & & & \\
y_{1}^{n-2} & y_{2}^{n-2} & & & & \\
\vdots & \vdots & & & & \\
y_{1}^{1} & y_{2}^{1} & y_{3}^{1} & \ldots & y_{n-1}^{1} \\
x_{1} & x_{2} & x_{3} & \ldots & x_{n-1} & x_{n}
\end{IEEEeqnarraybox*}
\end{equation}
We will consider that an \textit{edk} provides a solution to a certain problem if a constancy is found in its upper rows. We define constancy as the existence of $k$ levels of zeros, in the case of an \textit{edk} built by differences, or $k$ levels of ones, if it is built by divisions, with $k \in \{1,2,...,n-1\}$. In other words, an \textit{edk} is a solution if $\forall j \geq k$ and $\forall i \in \{1,2,...,n-j\}$ it is true that $|y_{i}^{j}|<\epsilon$ in first case, or $|y_{i}^{j}-1|<\epsilon$ in the second case, where $\epsilon=e^{r}$, $r\in\mathbb{R}_{*}^{-}$ is a preset and very small parameter.

%%%%%%%%%%%%%%%%%%%%%%%%%%%%%%%%%%%%%%%%%%%%%%%%%%%%%%%%%%%%%%%%%%%%%%%%%%%%%%%%%%%%%%%%%%%%%%%%%%%%%%%%%%%%%%

\subsection{The basic algorithms or \textit{kitas}} 

The \textit{kitas} are the set of algorithms that are applied in different configurations on a series or on an \textit{edk}. A \textit{kita} is the computational implementation, based on mathematical foundations, of a particular way of approaching the problem. Their combination allows us to find patterns of increasing complexity, while eventually reaching a solution.

\textbf{Basic (\textit{BAS}) and Divisions (\textit{DIV})}. These are used to build an \textit{edk} through differences or divisions, respectively.

\textbf{Reduction (\textit{RED})}. This \textit{kita} is used to focus on the upper levels of an \textit{edk} rather than its base. Let $edk_{n}$ be an \textit{edk} whose base $X$ is formed by $n$ elements. A reduction of $r\in\{1,2,...,n-2\}$ levels allows us to focus on the series:
\begin{equation}
    RED(r)\left[edk_{n}\right] \rightarrow \left\{y_{1}^{r}, y_{2}^{r}, ..., y_{n-r}^{r} \right\}
\end{equation}
\textit{BAS} is applied on this series to build a new \textit{edk}.

\textbf{Multi-Level (\textit{ML})}. This looks for patterns on the diagonals of the \textit{edk}. Let $dy, dx \in \mathbb{N}$ be the increments on the vertical and horizontal axis, respectively, starting from $edk_{n}$, a set of $dy+dx$ child \textit{edks} is generated:
\begin{equation}
    ML(dy, dx)\left[edk_{n}\right] \rightarrow \left\{edk_{m_{1}}, edk_{m_{2}}, ..., edk_{m_{dy+dx}}\right\}
\end{equation}
where $M=\left\{m_{1}, m_{2}, ..., m_{dy+dx}\right\}$, $m\in\mathbb{N}\hspace{0.1cm}|\hspace{0.1cm}2\leq m \leq n$ contains the sizes of each of the descendant \textit{edks}. The bases of the new child \textit{edks} are obtained by starting from the origin points $\left(p_{y_{0}}, p_{x_{0}}\right) = (0,1)$, $(1,1)$, $(2,1)$, and so on, following the left-hand outer face of the $edk_{n}$. From the corresponding origin, the elements located on the slope formed by $dy$ and $dx$ are extracted, applying the equation:
\begin{equation}
    \left(p_{y_{\alpha}}, p_{x_{\alpha}}\right) = (p_{y_{0}}+\alpha \cdot  dy, p_{x_{0}}+\alpha \cdot dx)
\label{eqml}
\end{equation}
with $\alpha$ taking values from 0, 1, 2, ... until the iteration corresponding to the limit of the \textit{edk}.

\textbf{Focusing (\textit{FOC})}. This is used to divide the original series into several subseries to find the implicit pattern in each. It uses two parameters, an initial offset $s$ and a list of the size of the fragments to be extracted $D=\left\{d_{1}, d_{2}, ..., d_{l}\right\}$ iteratively. For an original series $X$ of length $n$, if we denote the sum of divisions as $s_{d}= \sum_{r=1}^{l}d_{r}$, then it must be satisfied that $s\in \mathbb{N} \hspace{0.1cm} | \hspace{0.1cm} 0\leq s \leq s_{d}$ and $d, l \in \mathbb{N} \hspace{0.1cm} | \hspace{0.1cm} s+s_{d} \leq n \geq 4$. If this is fulfilled, then we start the \textit{kita} by moving the first $s$ elements of $X$, which are accumulated in what we will call \textit{remaining series} $R$. Next, $l$ new series are formed from successive iterations of fragments of size $d_{l}$ of the remaining \textit{shifted series}. The coordinates of the elements that form any series $S_m$, with $m\in \left\{1, 2, ..., l\right\}$, can be calculated using this equation, pruning the result where $c\in C_{S_{m}}>n$:
\begin{equation}
    C_{S_{m}} =\bigcup_{p=0}^{\lceil\frac{n-s}{s_{d}}\rceil-1}\bigcup_{q=1}^{d_{m}}\left\{s+f(m)+q+p\cdot s_{d}\right\}
\label{eqfoc}
\end{equation}
where $C_{S_{m}}$ is the set of coordinates, $\cup$ is the orderly union with preserving of all the elements of the series, and $f(m) = \sum_{r=1}^{m-1}d_{r}$ if $m \neq 1$, or $f(m) = 0$ otherwise.

Next, new fragments will be added to each subseries $S_{m}$, this time obtained from the $\textit{remaining series}$. These elements are inserted starting at the end of $R$ and starting with the last series $S_l$. On each of the $l$ iterations, we obtain the subseries of remainders $R_{m}$ that would correspond to $S_{m}$ extracting the last $d_m$ elements of $R$. Note that some $R_{m}$ series may have fewer elements than $d_m$ or may even be empty sequences if $R$ is exhausted. Therefore, focusing gives the series of the following type:
\begin{equation}
    FOC(s,D)\left[edk_{n}\right] \rightarrow X_{1}, ..., X_{l} = R_{1} \cup S_{1}, ..., R_{l} \cup S_{l}
\end{equation}
Finally, \textit{BAS} is applied on the series formed.

\textbf{Analogy (\textit{ANA})}. As in the case of \textit{FOC}, this \textit{kita} breaks the original series into new ones. It also uses two parameters: an initial offset $s \in \mathbb{N} \hspace{0.1cm} | \hspace{0.1cm} s \leq n-4$, which determines the number of elements ignored at the beginning of the series, and the number of elements per group or new child series $e \in \mathbb {N} \hspace{0.1cm} |\hspace{0.1cm} 1 < e \leq n-s-1 \hspace{0.1cm} , \hspace{0.1cm} (n-s)\%e \neq 0$. Then, the original series is divided into $t=\lceil(n-s)/e\rceil$ groups $G$ of size $e$; except for the last one ($G_{t}$), which must necessarily have a smaller size. Once the previous child series have been generated, for the first $t-1$ groups, a constancy is sought in the same row $r$ of their \textit{edk}, where $r \in \{1,2,...,n-1\}$. The prediction is made directly by placing the corresponding part of the \textit{edks} of the previous groups on the group $t$.

\textbf{Exponentiation (\textit{EXP})}. This is used to raise the elements of the base of an \textit{edk} to a certain power. In cases such as series formed by square numbers or any other power of $x$, this allows us to transform the series into a simpler one. Generally, by applying a exponentiation with exponent $e$ based on $edk_{n}$, where $e\in\mathbb{R}_{*}$ except in cases like $0 \in X$ or $\exists \hspace{0.1cm} x \in X \hspace{0.1cm} | \hspace{0.1cm} x < 0$, we obtain:
\begin{equation}
    EXP(e)\left[edk_{n}\right] \rightarrow \left\{x_{1}^{e}, x_{2}^{e}, ..., x_{n}^{e}\right\}
\end{equation}
The \textit{kita} ends by applying \textit{BAS} to the result.

\textbf{Logarithm (\textit{LOG})}. This calculates the first level of the \textit{edk} as the logarithm of the lower level, where the absolute value of each element is the base of the logarithm and the absolute value of the next element is the exponent:
\setlength{\arraycolsep}{0.0em}
\begin{eqnarray}
    & LOG\left[edk_{n}\right] \rightarrow \\ 
    & \left\{\log_{|x_{1}|}|x_{2}|, \log_{|x_{2}|}|x_{3}|, ..., \log_{|x_{n-1}|}|x_{n}| \right\} \nonumber
\end{eqnarray}
\setlength{\arraycolsep}{5pt}
The general equation that implements this \textit{kita} is:
\begin{equation}
    y_{i}^{1} = \log_{|x_{i}|}|x_{i+1}| 
\end{equation}
where $i \in \{1,2,...,n-1\}$. Finally, a new \textit{edk} is created using differences with this new sequence as the basis.

\textbf{Double Operation (\textit{DOP})}. This is used to build the next level to the base of an \textit{edk} by alternating two of the four basic arithmetic operations on the elements of the base. This allows us to obtain the pattern of a series in which there is a two-to-two arithmetic relationship between its elements. Assuming two operators $\mathcal{O}_{1}, \mathcal{O}_{2} \in \{+,-,\times,\div\}$, we have:
\setlength{\arraycolsep}{0.0em}
\begin{eqnarray}
    &DOP\left(\mathcal{O}_{1}, \mathcal{O}_{2}\right)\left[edk_{n}\right] \rightarrow \\ 
    & \left\{\mathcal{O}_{1}\left[x_{2},x_{1}\right], \mathcal{O}_{2}\left[x_{3},x_{2}\right], ..., \mathcal{O}_{j}\left[x_{n},x_{n-1}\right] \right\} \nonumber
\end{eqnarray}
\setlength{\arraycolsep}{5pt}
With $j = 1$ if $n$ is even, and $j = 2$ otherwise. In this way, the equation that governs the process is:
\begin{equation}
    \setlength{\nulldelimiterspace}{0pt}
    y_{i}^{1} = \mathcal{O}_{f(i)}\left[x_{i+1},x_{i}\right], \hspace{0.2cm} f(i) = \left\{\begin{IEEEeqnarraybox}[\relax][c]{l's}
    1, & $i\%2 \neq 0$\\
    2, & $i\%2 = 0$
    \end{IEEEeqnarraybox}\right.
\end{equation}
As in other cases, the \textit{kita} ends up by building an \textit{edk} by means of differences with the result series.

\textbf{Specular Symmetry (\textit{SSYM})}. This is used to predict the following items on the original series if a symmetric arrangement is detected. Generally, if from a position $j \in \{\round(n/2),...,n-1\}$ of the series, then its elements begin to repeat in reverse order and the remaining elements of the series can be obtained by continuing this subsequence.

\textbf{Repetition Symmetry (\textit{RSYM})}. This predicts the next elements in the basis of the \textit{edk} if a repeating group is detected throughout it. The series is divided into smaller groups, all of them of the same length, except for the last one. Different sizes of groups are tested until a symmetry is found. Once a repeating group is found, the elements to be predicted are obtained automatically. That is, let $G=\left\{x_{1}, x_{2}, ..., x_{j}\right\}$, $j \in \{2,3,...,n-1\}$ be the repeating group, the series can be extended indefinitely, according to the following equation:
\begin{equation}
    RSYM\left[edk_{n}\right] \rightarrow \{x_{1}, x_{2}, ..., x_{j}, x_{1}, x_{2}, ..., x_{j}, ...\}
\end{equation}
\textbf{Different Groups of Equal Elements (\textit{DGE})}. This groups consecutive and equal elements of the original series creating two new series: one of them formed by the element that is repeated in each group, and another formed with the sizes of the groups. That is, let $G=\left\{G_{1}, G_{2}, ..., G_{t}\right\}$ be the groups formed, with $t\in \mathbb{N} \hspace{0.1cm} | \hspace{0.1cm} 2\leq t \leq n-1$, the two new sequences $X_{1}$ and $X_{2}$ are formed as follows:
\begin{equation}
    \setlength{\nulldelimiterspace}{0pt}
    DGE\left[edk_{n}\right] \rightarrow \left\{\begin{IEEEeqnarraybox}[\relax][c]{l's}
    X_{1} = \left\{G_{1}\left[1\right], G_{2}\left[1\right], ..., G_{t}\left[1\right] \right\}\\
    X_{2} = \left\{\mathcal{L}\left[G_{1}\right], \mathcal{L}\left[G_{2}\right], ..., \mathcal{L}\left[G_{t}\right]\right\}
    \end{IEEEeqnarraybox}\right.
\end{equation}
where $G_{j}\left[1\right]$, $j\in\{1,2,...,t\}$ is the first element of the group $j$, and $\mathcal{L}$ is an operator which calculates its length. On these two series, two \textit{edks} are built by means of differences.

\textbf{Different Groups of Different Elements (\textit{DGD})}. This \textit{kita} is similar to $DGE$ but in this case the elements within each formed group are not equal. It also creates two subseries $X_{1}$ and $X_{2}$ from the basis of \textit{edk}: the first consisting of the groups of elements whose progression repeats in the sequence, and the second consisting of the lengths of those groups. The original series $X$ is divided into:
\begin{equation}
    \setlength{\nulldelimiterspace}{0pt}
    DGD\left[edk_{n}\right] \rightarrow \left\{\begin{IEEEeqnarraybox}[\relax][c]{l's}
    X_{1} = X_{\neq}\\
    X_{2} = \left\{\mathcal{L}\left[G_{1}\right], \mathcal{L}\left[G_{2}\right], ..., \mathcal{L}\left[G_{t}\right]\right\}
    \end{IEEEeqnarraybox}\right.
\label{eqDGD}
\end{equation}
where $X_{\neq}$ is a subsequence of $X$ with its different elements. Again, two new \textit{edks} are built on these series.

\textbf{Split Of Elements (\textit{SOE})}. This is used when in each element of $X$ all its digits have the same numerical value and there is a relationship between the number of digits of each element. In this way, two new subseries are obtained: one with the digit that is repeated in each element, and another with the number of digits of each of them:
\begin{equation}
    \setlength{\nulldelimiterspace}{0pt}
    SOE\left[edk_{n}\right] \rightarrow \left\{\begin{IEEEeqnarraybox}[\relax][c]{l's}
    X_{1} = \left\{\mathcal{R}\left[x_{1}\right], \mathcal{R}\left[x_{2}\right], ..., \mathcal{R}\left[x_{n}\right]\right\}\\
    X_{2} = \left\{\mathcal{N}\left[x_{1}\right], \mathcal{N}\left[x_{2}\right], ..., \mathcal{N}\left[x_{n}\right]\right\}
    \end{IEEEeqnarraybox}\right.
\end{equation}
where $\mathcal{R}\left[x_{i}\right]$ is an operator that selects the numeric value of the element $x_{i}$ of $X$, with $i\in \{1 ,2, ...,n\}$, and $\mathcal{N}\left[x_{i}\right]$ is another operator that calculates the number of digits in the element $x_{i}$. Finally, \textit{BAS} is applied on $X_{1}$ and $X_{2}$.

%%%%%%%%%%%%%%%%%%%%%%%%%%%%%%%%%%%%%%%%%%%%%%%%%%%%%%%%%%%%%%%%%%%%%%%%%%%%%%%%%%%%%%%%%%%%%%%%%%%%%%%%%%%%%%

\subsection{Search algorithms}

The pattern search method used by KitBit is based on a state tree. A connected tree or graph without cycles \cite{PEL} $G(S, K)$ is defined by a set of actions or \textit{kitas} $K=\left\{k_{1},k_{2}, ..., k_{n}\right\}$, $n\in\mathbb{N}$ that form the axes of the graph, and a set of states $S=\left\{s_{1},s_{2}, ..., s_{m}\right\}$ that form its vertices, where $m \in \mathbb{N} \hspace{0.1cm} | \hspace{0.1cm} 1 \leq m \leq 1+\sum_{r=1}^{depth-1} n^{r}$ and $depth\in \mathbb{N} \hspace{0.1cm} | \hspace{0.1cm} depth \geq 1$ is the depth of the tree. Aside from the root state, the other nodes have a single parent and a number of descendants ranging from 0 to $n$ that are visually identified as vertices connected to the same parent node. Any state $s \in S$ is defined as a four-components structure, according to (\ref{eqsa}).

\begin{equation}
    s=\left\{\begin{array}{c}
        \left\{edk_{1}, edk_{2}, ..., edk_{l}\right\} \\
        k \in K \\
        \left\{c_{1}, c_{2}, ..., c_{l}\right\} \hspace{0.1cm} , \hspace{0.1cm} c \in \{0,1\} \\
        \left\{f_{1}, f_{2}, ..., f_{l}\right\} \hspace{0.1cm} , \hspace{0.1cm} f \in \{1,2,...,l\}
    \end{array}\right\}
\label{eqsa}
\end{equation}
where the first component is an ordered sequence of $l\in \mathbb{N}$ \textit{edks} on which a $k$ action has been applied. The second component is the action. The third component is an ordered series of zeros and ones, in which a one at position $i \in \{1,2,...,l\}$ indicates that the corresponding \textit{edk} is a solution (it has a constancy of zeros or ones in its top rows) and a zero that it is not. The fourth component is another ordered series of natural numbers that indicate the position of the parent of the \textit{edk} $i$ in the list of \textit{edks} of the previous state. When all of the elements are ones in the third component of $s$, then this means that a regularity has been found in the data and this state constitutes a solution to the problem.

This tree structure can be traversed in several ways by implementing different pattern search strategies. KitBit uses an uninformed search algorithm, with the popular Breadth First Search (BFS) and Depth First Search (DFS) strategies. In both algorithms, reaching a solution state also implies that we obtain the path of actions performed in each previous state. We denote this list as $K_{s} \subseteq K$, where $\subseteq$ is an orderly subset of $K$ with preserving of all the elements. This sequence of kitas constitutes a description of the pattern inherent in the numerical series, compressing the series into a
recurrence formula, although not in the mathematical sense but as a set of algorithms that, when executed, produce the numerical series in question. Therefore, KitBit is capable of both predicting new items and providing the actions that generate them. The number of new items generated by $K_{s}$ may be unlimited or not, depending on the structure of the series and the \textit{kitas} used. Likewise, there may be more than one list of actions $K_{s}$ solution to the problem. Therefore, KitBit implements the possibility of not stopping when a first solution is found but continuing to generate new states to search for all solutions. Finally, following the Minimum Description Length (MDL) principle \cite{GRUN}, we will consider the optimal model as the simplest, this being the one that requires the least number of \textit{kitas} to predict the greatest number of new elements sequentially.

%%%%%%%%%%%%%%%%%%%%%%%%%%%%%%%%%%%%%%%%%%%%%%%%%%%%%%%%%%%%%%%%%%%%%%%%%%%%%%%%%%%%%%%%%%%%%%%%%%%%%%%%%%%%%%

\subsection{Prediction}

Once the sequence of actions to solve the problem $K_{s}$ has been determined, they are applied in reverse order to obtain new elements of the series. Starting from the final state reached, and for all \textit{edks} contained in the solution state path, the operations to be carried out are the following.

Firstly, if the \textit{edk} is a solution, we add as many zeros or ones at the top of it as elements to be predicted, depending on whether the constancy was achieved using \textit{BAS} or \textit{DIV}, respectively. These elements are denoted as $\beta\in B$ and are said to warp the \textit{edk}, distorting its initial triangular shape. Next, the \textit{edk} is traversed in an inverse way by applying (\ref{eqpred}).
\begin{equation}
    \setlength{\nulldelimiterspace}{0pt}
    \beta_{i}^{j} = \left\{\begin{IEEEeqnarraybox}[\relax][c]{l's}
    \beta_{i}^{j+1}+\beta_{i-1}^{j}, & BAS \\
    \beta_{i}^{j+1}\cdot\beta_{i-1}^{j}, & DIV
    \end{IEEEeqnarraybox}\right.
\label{eqpred}
\end{equation}
where $i \in \left\{1,2,...,n_{\beta_{j+1}}\right\}$, $j \in \left\{0,1,...,n_{j}-1\right\}$, $\beta_{0}^{0}=x_n$ and $\beta_{0}^{j}=y_{n-j}^{j}$, $\forall j \neq 0$. $n_{\beta_{j+1}}$ is the number of deformation elements that exist in row $j+1$ and $n_j$ is the number of rows involved. The prediction process would start with $\beta_{i}^{j+1}=\beta_{\min(i)}^{\max(j)}$ as the point of origin; that is, with the deformation element located at the highest level and furthest to the left-hand. In general, when this process reaches the base of the \textit{edk}, the predicted terms are inserted directly into the base of its parent \textit{edk}, except for \textit{RED}, \textit{LOG}, \textit{DOP}, \textit{ML} and \textit{FOC}. In \textit{RED}, the elements are inserted into the index row $r$ of the parent \textit{edk}. For \textit{LOG} and \textit{DOP}, the predicted numbers are placed in the $r=1$ row. Meanwhile, for \textit{ML} these new terms are placed at the slopes of the parent \textit{edk}, using (\ref{eqml}) to calculate the coordinates. Finally, for \textit{FOC}, the insertion occurs at the base of the \textit{edk} but computing the new coordinates with (\ref{eqfoc}), changing $n$ to $n+n_{\beta}$, where $n_{\beta}$ is the number of elements to predict.

Next, it is necessary to compute the inverse function of the \textit{kita} on the parent \textit{edk}. For \textit{EXP} it implies to apply the inverse of the exponent that was passed as an argument (that is, $1/e$). In the case of \textit{LOG}, the inverse action works on the basis of the \textit{edk} and its upper row, using the equation:
\begin{equation}
    \beta_i^{0} = \left(\beta_{i-1}^{0}\right)^{\beta_{i}^{1}}
\end{equation}
Similarly, in \textit{DOP} this step consists of applying the opposite operator to $\mathcal{O}_{1}$ and $\mathcal{O}_{2}$ between the base and its top level:
\begin{equation}
\setlength{\nulldelimiterspace}{0pt}
    \beta_{i}^{0} = \mathcal{O}'_{f(g)}\left[\beta_{i}^{1},\beta_{i-1}^{0}\right], \hspace{0.2cm} f(g) = \left\{\begin{IEEEeqnarraybox}[\relax][c]{l's}
    1, & $g(i,n)$\\
    2, & $\overline{g(i,n)}$
    \end{IEEEeqnarraybox}\right.
\end{equation}
where $\mathcal{O}'_{1}$ is the opposite operation of $\mathcal{O}_{1}$, and $g(i,n) = (i\%2 \neq 0 \wedge n\%2 \neq 0) \vee (i\%2 = 0 \wedge n\%2 = 0)$. Finally, in \textit{DGE}, \textit{DGD} and \textit{SOE}, if we denote the two child series of the parent \textit{edk} under the indices $a$ for the first, and $b$ for the second, then the respective base element of the parent \textit{edk} $\beta^{0}\in B$ is obtained for each pair of predicted elements $\beta_{a}^{0}\in B_{a}$, $\beta_{a}^{0}\in B_{b}$ located at the same positions in both secondary series. In \textit{DGE} the element $\beta_{a}^{0}$ is repeated $\beta_{b}^{0}$ times. In \textit{DGD}, the next term is formed by extracting as many items as $\beta_{b}^{0}$ indicates from $X_{1}$. In \textit{SOE}, $\beta^{0}$ is constructed as $\beta^{0} = \beta_{a}^{0}\cdot\left(10^{0}+ 10^ {1}+...+10^{\beta_{b}^{0}}\right)$. 

This process is applied repeatedly and propagates until the last state is reached.

Table \ref{tabex1} shows some example applications of the \textit{kitas}. For clarity, very simple series that resolve to a single \textit{kita} have been selected. And for reasons of space, we don't show all the terms that could be predicted by applying our method. In each row, the original series appears at the base of the \textit{edk} in the `Original series/\textit{edk}' column or as a single series. The `\textit{kita} \textit{edks}' and `\textit{kita} prediction' columns contain the \textit{edk} generated by the \textit{kita} and the prediction made, respectively. In them we can also see the achievement of a regularity by applying the \textit{kitas}. Finally, the `Result series/\textit{edk} ' column contains the final prediction.

\begin{table*}[!t]
\renewcommand{\arraystretch}{1.3}
\caption{\textit{Kita} Application Examples.}
\label{tabex1}
\centering
\begin{tabular}{c|c|c|c|c}
\hline
\textit{kita} & Original series/\textit{edk} & \textit{kita} \textit{edks} & \textit{kita} prediction & Result series/\textit{edk}\\ \hline
\textit{RED}(1) & $\begin{IEEEeqnarraybox*}[][c]{,r/r/r/r/r,}
& & & & \\ \noalign{\vspace{-0.2ex}}
\vdots & \vdots & \vdots & & \\ \noalign{\vspace{-0.11ex}}
1 & 2 & 3 & 4 & \\ \noalign{\vspace{-0.11ex}}
3 & 3 & 6 & 18 & 72 \\ \noalign{\vspace{-1.5ex}}
& & & & 
\end{IEEEeqnarraybox*}$ & 
$\begin{IEEEeqnarraybox*}[][c]{,r/r/r/r,}
0 & 0 & & \\ \noalign{\vspace{-0.11ex}}
1 & 1 & 1 & \\ \noalign{\vspace{-0.11ex}}
1 & 2 & 3 & 4
\end{IEEEeqnarraybox*}$ & $\begin{IEEEeqnarraybox*}[][c]{,r/r/r/r/r/r,}
0 & 0 & 0 & 0 & & \\ \noalign{\vspace{-0.11ex}}
1 & 1 & 1 & 1 & 1 & \\ \noalign{\vspace{-0.11ex}}
1 & 2 & 3 & 4 & 5 & 6
\end{IEEEeqnarraybox*}$ & $\begin{IEEEeqnarraybox*}[][c]{,r/r/r/r/r/r/r/r/r,}
& & & & & & \\ \noalign{\vspace{-0.2ex}}
1 & 2 & 3 & 4 & 5 & 6 \\ \noalign{\vspace{-0.11ex}}
3 & 3 & 6 & 18 & 72 & 360 & 2160 \\ \noalign{\vspace{-1.5ex}}
& & & & & &
\end{IEEEeqnarraybox*}$ \\ \hline
%%%%%%%%%%%%%%%%%%%%%%%%%%%%%%%%%%%%%%%%%%%%%%%%%%%%%%%%%%%%%%%%%%%%%%%%%%%%%%%%%%%%%%%%%%%%%%%%%%%%%%%%%%%%%%%%%%%%%%%%%
\textit{ML}(1,1) & $\begin{IEEEeqnarraybox*}[][c]{,r/r/r/r/r/r,}
& & & & & \\ \noalign{\vspace{-0.2ex}}
\vdots & \vdots & 2 & & & \\ \noalign{\vspace{-0.11ex}}
 & 2 & 3 & 5 & & \\ \noalign{\vspace{-0.11ex}}
2 & 3 & 5 & 8 & 13 & \\ \noalign{\vspace{-0.11ex}}
3 & 5 & 8 & 13 & 21 & 34 \\ \noalign{\vspace{-1.5ex}}
& & & & &
\end{IEEEeqnarraybox*}$ & 
$\begin{IEEEeqnarraybox*}[][c]{,r/r/r/r/r/r/r,}
& & & & & & \\ \noalign{\vspace{-0.2ex}}
0 & 0 & & & 0 & 0 & \\ \noalign{\vspace{-0.11ex}}
3 & 3 & 3 & & 2 & 2 & 2 \\ \noalign{\vspace{-1.5ex}}
& & & & & & 
\end{IEEEeqnarraybox*}$ & 
$\begin{IEEEeqnarraybox*}[][c]{,r/r/r/r/r/r/r/r/r,}
& & & & & & & & \\ \noalign{\vspace{-0.2ex}}
0 & 0 & 0 & & & 0 & 0 & 0 & \\ \noalign{\vspace{-0.11ex}}
3 & 3 & 3 & 3 & & 2 & 2 & 2 & 2 \\ \noalign{\vspace{-1.5ex}}
& & & & & & & & 
\end{IEEEeqnarraybox*}$ & 
$\begin{IEEEeqnarraybox*}[][c]{,r/r/r/r/r/r/r/r,}
& & & & & & & \\ \noalign{\vspace{-0.2ex}}
 & & & 2 & & & & \\ \noalign{\vspace{-0.11ex}}
 & & 2 & 3 & 5 & & & \\ \noalign{\vspace{-0.11ex}}
 & 2 & 3 & 5 & 8 & 13 & & \\ \noalign{\vspace{-0.11ex}}
2 & 3 & 5 & 8 & 13 & 21 & 34 & \\ \noalign{\vspace{-0.11ex}}
3 & 5 & 8 & 13 & 21 & 34 & 55 & 89 \\ \noalign{\vspace{-1.5ex}}
& & & & & & &
\end{IEEEeqnarraybox*}$ \\ \hline
%%%%%%%%%%%%%%%%%%%%%%%%%%%%%%%%%%%%%%%%%%%%%%%%%%%%%%%%%%%%%%%%%%%%%%%%%%%%%%%%%%%%%%%%%%%%%%%%%%%%%%%%%%%%%%%%%%%%%%%%%
\textit{FOC}(0,\{1,1\}) & $\{-6, -4, 0, 2, 6, 8\}$ & 
$\begin{IEEEeqnarraybox*}[][c]{,r/r/r/r/r/r/r,}
& & & & & & \\ \noalign{\vspace{-0.2ex}}
0 & & & & 0 & & \\ \noalign{\vspace{-0.11ex}}
6 & 6 & & & 6 & 6 & \\ \noalign{\vspace{-0.11ex}}
-6 & 0 & 6 & & -4 & 2 & 8\\ \noalign{\vspace{-1.5ex}}
& & & & & &
\end{IEEEeqnarraybox*}$ & 
$\begin{IEEEeqnarraybox*}[][c]{,r/r/r/r/r/r/r/r/r,}
& & & & & & & & \\ \noalign{\vspace{-0.2ex}}
0 & 0 & & & & 0 & 0 & & \\ \noalign{\vspace{-0.11ex}}
6 & 6 & 6 & & & 6 & 6 & 6 & \\ \noalign{\vspace{-0.11ex}}
-6 & 0 & 6 & 12 & & -4 & 2 & 8 & 14 \\ \noalign{\vspace{-1.5ex}}
& & & & & & & & 
\end{IEEEeqnarraybox*}$ & $\{-6, -4, 0, 2, 6, 8, 12, 14\}$ \\ \hline 
%%%%%%%%%%%%%%%%%%%%%%%%%%%%%%%%%%%%%%%%%%%%%%%%%%%%%%%%%%%%%%%%%%%%%%%%%%%%%%%%%%%%%%%%%%%%%%%%%%%%%%%%%%%%%%%%%%%%%%%%%
\textit{ANA}(0,4) & $\{1, 3, 5, 7, 2, 4\}$ & 
$\begin{IEEEeqnarraybox*}[][c]{,r/r/r/r/r/r/r,}
& & & & & & \\ \noalign{\vspace{-0.2ex}}
0 & 0 & & & & & \\ \noalign{\vspace{-0.11ex}}
2 & 2 & 2 & & & 2 & \\ \noalign{\vspace{-0.11ex}}
1 & 3 & 5 & 7 & & 2 & 4\\ \noalign{\vspace{-0.11ex}}
\end{IEEEeqnarraybox*}$ & 
$\begin{IEEEeqnarraybox*}[][c]{,r/r/r/r/r/r/r/r/r,}
& & & & & & & & \\ \noalign{\vspace{-0.2ex}}
0 & 0 & & & & 0 & 0 & & \\ \noalign{\vspace{-0.11ex}}
2 & 2 & 2 & & & 2 & 2 & 2 &  \\ \noalign{\vspace{-0.11ex}}
1 & 3 & 5 & 7 & & 2 & 4 & 6 & 8 \\ \noalign{\vspace{-1.5ex}}
& & & & & & & & 
\end{IEEEeqnarraybox*}$ & $\{1, 3, 5, 7, 2, 4, 6, 8\}$ \\ \hline
%%%%%%%%%%%%%%%%%%%%%%%%%%%%%%%%%%%%%%%%%%%%%%%%%%%%%%%%%%%%%%%%%%%%%%%%%%%%%%%%%%%%%%%%%%%%%%%%%%%%%%%%%%%%%%%%%%%%%%%%%
\textit{EXP}(1/4) & $\{1, 16, 81, 256\}$ & 
$\begin{IEEEeqnarraybox*}[][c]{,r/r/r/r,}
& & & \\ \noalign{\vspace{-0.2ex}}
0 & 0 & & \\ \noalign{\vspace{-0.11ex}}
1 & 1 & 1 & \\ \noalign{\vspace{-0.11ex}}
1 & 2 & 3 & 4 \\ \noalign{\vspace{-1.5ex}}
& & &
\end{IEEEeqnarraybox*}$ & $\begin{IEEEeqnarraybox*}[][c]{,r/r/r/r/r/r,}
& & & & & \\ \noalign{\vspace{-0.2ex}}
0 & 0 & 0 & 0 & & \\ \noalign{\vspace{-0.11ex}}
1 & 1 & 1 & 1 & 1 & \\ \noalign{\vspace{-0.11ex}}
1 & 2 & 3 & 4 & 5 & 6 \\ \noalign{\vspace{-1.5ex}}
& & & & & 
\end{IEEEeqnarraybox*}$ & $\{1, 16, 81, 256, 625, 1296\}$ \\ \hline 
%%%%%%%%%%%%%%%%%%%%%%%%%%%%%%%%%%%%%%%%%%%%%%%%%%%%%%%%%%%%%%%%%%%%%%%%%%%%%%%%%%%%%%%%%%%%%%%%%%%%%%%%%%%%%%%%%%%%%%%%%%%%%%%
\textit{LOG} & $\{65536, 256, 16, 4\}$ & 
$\begin{IEEEeqnarraybox*}[][c]{,l/l/l,}
& & \\ \noalign{\vspace{-0.2ex}}
0 & 0 & \\ \noalign{\vspace{-0.11ex}}
0.5 & 0.5 & 0.5 \\ \noalign{\vspace{-1.5ex}}
& &
\end{IEEEeqnarraybox*}$ & $\begin{IEEEeqnarraybox*}[][c]{,l/l/l/l/l,}
& & & & \\ \noalign{\vspace{-0.2ex}}
0 & 0 & 0 & 0 & \\ \noalign{\vspace{-0.11ex}}
0.5 & 0.5 & 0.5 & 0.5 & 0.5 \\ \noalign{\vspace{-1.5ex}}
& & & &
\end{IEEEeqnarraybox*}$ & $\{65536, 256, 16, 4, 2, \sqrt{2}\}$ \\ \hline 
%%%%%%%%%%%%%%%%%%%%%%%%%%%%%%%%%%%%%%%%%%%%%%%%%%%%%%%%%%%%%%%%%%%%%%%%%%%%%%%%%%%%%%%%%%%%%%%%%%%%%%%%%%%%%%%%%%%%%%%%%
\textit{DOP}($-$,$\div$) & $\{3, 5, 10, 12, 24\}$ & 
$\begin{IEEEeqnarraybox*}[][c]{,r/r/r/r,}
& & & \\ \noalign{\vspace{-0.2ex}}
0 & 0 & 0 & \\ \noalign{\vspace{-0.11ex}}
2 & 2 & 2 & 2 \\ \noalign{\vspace{-1.5ex}}
& & &
\end{IEEEeqnarraybox*}$ & $\begin{IEEEeqnarraybox*}[][c]{,r/r/r/r/r/r,}
& & & & & \\ \noalign{\vspace{-0.2ex}}
0 & 0 & 0 & 0 & 0 & \\ \noalign{\vspace{-0.11ex}}
2 & 2 & 2 & 2 & 2 & 2 \\ \noalign{\vspace{-1.5ex}}
& & & & &
\end{IEEEeqnarraybox*}$ & $\{3, 5, 10, 12, 24, 26, 52\}$ \\ \hline 
%%%%%%%%%%%%%%%%%%%%%%%%%%%%%%%%%%%%%%%%%%%%%%%%%%%%%%%%%%%%%%%%%%%%%%%%%%%%%%%%%%%%%%%%%%%%%%%%%%%%%%%%%%%%%%%%%%%%%%%%%
\textit{SSYM} & $\{1,2,4,8,4\}$ & - & - & $\{1,2,4,8,4,2,1\}$ \\ \hline 
%%%%%%%%%%%%%%%%%%%%%%%%%%%%%%%%%%%%%%%%%%%%%%%%%%%%%%%%%%%%%%%%%%%%%%%%%%%%%%%%%%%%%%%%%%%%%%%%%%%%%%%%%%%%%%%%%%%%%%%%%
\textit{RSYM} & $\{1,0,2,1,0\}$ & - & - & $\{1,0,2,1,0,2,1\}$ \\ \hline 
%%%%%%%%%%%%%%%%%%%%%%%%%%%%%%%%%%%%%%%%%%%%%%%%%%%%%%%%%%%%%%%%%%%%%%%%%%%%%%%%%%%%%%%%%%%%%%%%%%%%%%%%%%%%%%%%%%%%%%%%%
\textit{DGE} & $\{1, 3, 3, 5, 5, 5\}$ & 
$\begin{IEEEeqnarraybox*}[][c]{,r/r/r/r/r/r/r,}
& & & & & & \\ \noalign{\vspace{-0.2ex}}
0 & & & & 0 & & \\ \noalign{\vspace{-0.11ex}}
2 & 2 & & & 1 & 1 & \\ \noalign{\vspace{-0.11ex}}
1 & 3 & 5 & & 1 & 2 & 3 \\ \noalign{\vspace{-1.5ex}}
& & & & & & 
\end{IEEEeqnarraybox*}$ & 
$\begin{IEEEeqnarraybox*}[][c]{,r/r/r/r/r/r/r/r/r,}
& & & & & & & & \\ \noalign{\vspace{-0.2ex}}
0 & 0 & & & & 0 & 0 & & \\ \noalign{\vspace{-0.11ex}}
2 & 2 & 2 & & & 1 & 1 & 1 & \\ \noalign{\vspace{-0.11ex}}
1 & 3 & 5 & 7 & & 1 & 2 & 3 & 4 \\ \noalign{\vspace{-1.5ex}}
& & & & & & & & 
\end{IEEEeqnarraybox*}$ & $\{1, 3, 3, 5, 5, 5, 7, 7\}$ \\ \hline 
%%%%%%%%%%%%%%%%%%%%%%%%%%%%%%%%%%%%%%%%%%%%%%%%%%%%%%%%%%%%%%%%%%%%%%%%%%%%%%%%%%%%%%%%%%%%%%%%%%%%%%%%%%%%%%%%%%%%%%%%%
\textit{DGD} & $\{2, 2, 4, 2, 4, 6\}$ & 
$\begin{IEEEeqnarraybox*}[][c]{,r/r/r/r/r/r/r,}
& & & & & & \\ \noalign{\vspace{-0.2ex}}
0 & & & & 0 & & \\ \noalign{\vspace{-0.11ex}}
2 & 2 & & & 1 & 1 & \\ \noalign{\vspace{-0.11ex}}
2 & 4 & 6 & & 1 & 2 & 3 \\ \noalign{\vspace{-1.5ex}}
& & & & & & 
\end{IEEEeqnarraybox*}$ & 
$\begin{IEEEeqnarraybox*}[][c]{,r/r/r/r/r/r/r/r/r,}
& & & & & & & & \\ \noalign{\vspace{-0.2ex}}
0 & 0 & & & & 0 & 0 & &  \\ \noalign{\vspace{-0.11ex}}
2 & 2 & 2 & & & 1 & 1 & 1 & \\ \noalign{\vspace{-0.11ex}}
2 & 4 & 6 & 8 & & 1 & 2 & 3 & 4 \\ \noalign{\vspace{-1.5ex}}
& & & & & & & & 
\end{IEEEeqnarraybox*}$ & $\{2, 2, 4, 2, 4, 6, 2, 4\}$ \\ \hline
%%%%%%%%%%%%%%%%%%%%%%%%%%%%%%%%%%%%%%%%%%%%%%%%%%%%%%%%%%%%%%%%%%%%%%%%%%%%%%%%%%%%%%%%%%%%%%%%%%%%%%%%%%%%%%%%%%%%%%%%%
\textit{SOE} & $\{1, 11, 111\}$ & 
$\begin{IEEEeqnarraybox*}[][c]{,r/r/r/r/r/r/r,}
& & & & & & \\ \noalign{\vspace{-0.2ex}}
 & & & & 0 & & \\ \noalign{\vspace{-0.11ex}}
0 & 0 & & & 1 & 1 &  \\ \noalign{\vspace{-0.11ex}}
1 & 1 & 1 & & 1 & 2 & 3 \\ \noalign{\vspace{-1.5ex}}
& & & & & & 
\end{IEEEeqnarraybox*}$ & 
$\begin{IEEEeqnarraybox*}[][c]{,r/r/r/r/r/r/r/r/r/r/r,}
& & & & & & & & & &  \\ \noalign{\vspace{-0.2ex}}
& & & & & & 0 & 0 & 0 & & \\ \noalign{\vspace{-0.11ex}}
0 & 0 & 0 & 0 & & & 1 & 1 & 1 & 1 & \\ \noalign{\vspace{-0.11ex}}
1 & 1 & 1 & 1 & 1 & & 1 & 2 & 3 & 4 & 5 \\ \noalign{\vspace{-1.5ex}}
& & & & & & & & & & 
\end{IEEEeqnarraybox*}$ & $\{1, 11, 111, 1111, 11111\}$ \\ \hline
\end{tabular}
\end{table*}

%%%%%%%%%%%%%%%%%%%%%%%%%%%%%%%%%%%%%%%%%%%%%%%%%%%%%%%%%%%%%%%%%%%%%%%%%%%%%%%%%%%%%%%%%%%%%%%%%%%%%%%%%%%%%%%%%%%%%%%%%%%%%%%%%%%%%%%%%%%%%%%%%%%%%%%%%%%%%%%%%%%%%%%%%%%%%%%%%%%%%%%%%%%%%%%%%%%%%%%%%%%%%%%%%%%%%%%%%%%%

\section{Tests and results} 

An initial sample of 90 series, collected from real IQ tests, has been used to launch the model. Next, KitBit has been confronted with all the series used to test the computational models oriented to the resolution of numerical series found in the bibliography. Finally, the model faces the resolution of the series collected in the OEIS database, which is composed of more than 340,000 series of all kinds, without any type of limitation regarding its nature and characteristics. In all cases, the best model of those found is chosen based on the criteria of MDL explained in subsection 3.3.

\subsection{Series from IQ test} 

Initially, the system was challenged with the set of 90 non-trivial series based on the IQ tests that are summarized in Table \ref{tabIQ}. A computer with an Intel Core i7 processor and 16 GB of RAM running a 64-bit Windows 10 operating system was used. The results are shown in Tables \ref{tabIQres1} and \ref{tabIQres2}.

\begin{table*}[!t]
\renewcommand{\arraystretch}{1.3}
\caption{Number Series from IQ Test.}
\label{tabIQ}
\centering
\begin{tabular}{c p{5cm} c p{5cm} c p{5cm}}
\hline
i & sequence & i & sequence & i & sequence \\ \hline
0 & 0, 1, 1.7071, 2.3660, 3, 3.6180, 4.2247, 4.8229 & 30 & 2, 2, 4, 4, 8, 8, 16, 16, 32, 32, 64, 64, 128, 128, 256, 256 & 60 & 0, 2, 9, 28, 75, 186, 441, 1016, 2295, 5110, 11253, 24564 \\
1 & 2, 16, 4, 256, 16, 65536, 256, 4294967296, 65536 & 31 & 8, 6, 4, 3, 1, -1, -2, -4, -6, -7, -9, -11, -12, -14, -16 & 61 & 3, 6, 18, 36, 108, 216, 648, 1296, 3888, 7776, 23328, 46656, 139968 \\ 
2 & 3, 5, 8, 13, 21, 34, 55, 89, 144, 233, 377, 610, 987, 1597, 2584 & 32 & 362880, 40320, 5040, 720, 120, 24, 6, 2, 1, 1 & 62 & 2, 3, 5, 9, 17, 33, 65, 129, 257, 513, 1025, 2049, 4097, 8193 \\ 
3 & 1, 3, 7, 12, 18, 26, 35, 45, 57, 70, 84, 100, 117, 135 & 33 & 14, 1, -5.5, -8.75, -10.375, -11.1875, -11.59375 & 63 & 2, 10, 26, 50, 82, 122, 170, 226, 290, 362, 442, 530, 626 \\ 
4 & 0, 3, 12, 17, 102, 109, 872, 881 & 34 & -1, -1, 0, 2, 5, 9, 14, 20 & 64 & 4, 1, 0, 1, 4, 9, 16, 25, 36 \\ 
5 & 0, 3, 8, 24, 63, 168, 440, 1155, 3024, 7920 & 35 & -2, 3, 1, 4, 5, 9, 14, 23 & 65 & 9, 3, 6, 6, 2, 5, 3, 1, 4, 0, 0, 3, -3 \\
6 & 1, 3, 6, 10, 15, 21, 28, 36 & 36 & -3, -1, -4, 0, -5, 1, -6, 2, -7 & 66 & -9, 1, -5, 3, -4, 2, -6, -2, -11, -9 \\ 
7 & 1, 0, -1, 0, 1, 0, -1, 0, 1 & 37 & -4, -1, -3, 0, -2, 1, -1, 2, 0 & 67 & 97.5, 57, 30, 12, 0, -8, -13.333 \\ 
8 & 1, 3, 5, 7, 9, 11, 13, 15, 17 & 38 & 1, 2, 0, 2, -1, 2, -2, 2, -3 & 68 & 6, 4, 3, 3, 2, 2, 3, 1, 6, 0, 11, -1 \\ 
9 & 1, 2, 4, 7, 11, 16, 22, 29 & 39 & 1, -1, 0, -3, -1, -5, -2, -7, -3, -9 & 69 & 7, 16, 52, 196, 772, 3076, 12292 \\ 
10 & 1, 4, 9, 16, 25, 36, 49, 64, 81 & 40 & 0, 1, 0, -1, 0, 1, 0, -1, 0, 1, 0, -1 & 70 & 2, 3, 5, 4, 2, 4, 7, 5, 2, 5, 9, 6, 2, 6, 11, 7, 2 \\
11 & 3, 5, 10, 12, 24, 26, 52, 54, 108 & 41 & -2, 0, -3, 1, -4, 2, -5, 3, -6 & 71 & 8, 0, 2, 0, 0, 0, 2, 0, 8, 0, 18, 0 \\ 
12 & 3, 4, 8, 17, 33, 58, 94, 143 & 42 & 23, 34, 45, 56, 67, 78 & 72 & 3, 0, 1, -1, -2, -5, -9, -16 \\ 
13 & 3, 6, 18, 72, 360, 2160, 15120, 120960 & 43 & -4, -3, 0, 5, 12, 21, 32 & 73 & 1, 0, -1, -1, -2, -1, -3, -4, -1, -5 \\ 
14 & 4, 5, 8, 13, 20, 29, 40, 53 & 44 &-4, -2, 2, 8, 16, 26, 38 & 74 & 3, 9, 22.5, 45, 67.5, 67.5, 33.75, 0 \\ 
15 & 5, 11, 17, 23, 29, 35, 41 & 45 & 6, 4, 0, -6, -14, -24, -36 & 75 & 0, 2, 9, 24, 50, 90, 147, 224 \\
16 & 11, 9, 7, 5, 3, 1, -1, -3 & 46 & -2, 0, 4, 10, 18, 28, 40 & 76 & 3, 8, 16, 28, 45, 68, 98, 136 \\ 
17 & 30, 29, 27, 26, 24, 23, 21, 20, 18, 17, 15 & 47 & -6, -1, 5, 12, 20, 29, 39 & 77 & 0, 0, 4, 5, 14, 16, 30, 33, 52, 56 \\
18 & 144, 121, 100, 81, 64, 49, 36, 25 & 48 & 6400, 1600, 400, 100, 25, 6.25, 1.5625 & 78 & 0, 8, 15, 35, 48, 80, 99, 143, 168, 224 \\ 
19 & 2, 2, 4, 6, 10, 16, 26, 42, 68 & 49 & 0, 7, 24, 51, 88, 135, 192 & 79 & 4, 32, 108, 256, 500, 864, 1372, 2048 \\ 
20 & 81, 27, 9, 3, 1, 1/3, 1/9 & 50 & 2, 4, 12, 48, 240, 1440, 10080 & 80 & 0, 17, 74, 195, 404, 725, 1182, 1799 \\ 
21 & 1, 1, 2, 3, 5, 8, 13, 21 & 51 & -10, 12, 44, 86, 138, 200, 272 & 81 & 6, 24, 60, 120, 210, 336, 504, 720 \\ 
22 & 21, 20, 18, 15, 11, 6, 0 & 52 & 1, 0, 1, 1, 1, 2, 1, 3, 1, 4 & 82 & 3, 6, 15, 42, 123, 366, 1095, 3282 \\ 
23 & 8, 6, 7, 5, 6, 4, 5, 3, 4 & 53 & -1, 2, -2, -4, 8, -32, -256, 8192 & 83 & 23, 31, 53, 83, 135, 217, 351, 567 \\ 
24 & 4294967296, 65536, 256, 16, 4, 2 & 54 & -10, 0, 15, 35, 60, 90, 125, 165 & 84 & 2, 6, 19, 53, 126, 262, 491, 849, 1378 \\
25 & 3, 7, 14, 24, 37, 53, 72 & 55 & -2, -1, 1, 5, 13, 29, 61, 125 & 85 & 1, 2, 5, 9, 16, 27, 45, 74 \\ 
26 & -3, -1, 2, 6, 11, 17, 24 & 56 & -3, -2, 0, 1, 3, 4, 6, 7, 9 & 86 & 0, 2, 6, 12, 20, 30, 42, 56, 72, 90, 110, 132 \\
27 & -1, 0, 3, 8, 15, 24, 35 & 57 & 1, 2, 6, 14, 29, 56, 102, 176, 289 & 87 & 3, 5, 8, 12, 17, 23, 30, 38 \\
28 & -9, 2, 12, 21, 29, 36, 42 & 58 & 0, 1, 2, 8, 29, 80, 181, 357, 638 & 88 & 5, 8, 20, 68, 260, 1028, 4100 \\
29 & 1, 0, 0, 1, 3, 6, 10, 15 & 59 & 8, 1, 0, -1, -8, -27, -64, -125, -216 & 89 & -6, -4, 0, 2, 6, 8, 12, 14, 18 \\ \hline
\end{tabular}
\end{table*}

\begin{table*}[!t]
\renewcommand{\arraystretch}{1.3}
\caption{Results Obtained for the Series in Table \ref{tabIQ}.}
\label{tabIQres1}
\centering
\begin{tabular}{cccccccccccccccc}
\hline
& results (\%) & \multicolumn{4}{c}{depth of the graph} & time (ms) & \multicolumn{8}{c}{$n_e$ (\%)} \\ \cline{2-15}
& solved & 1 (\%) & 2 (\%) & 3 (\%) & 4 (\%) & $t_{med}$ & 3 & 4 & 5 & 6 & 7 & 8 & 9 & 10-16 \\ \hline 
\textit{S1Z} & 97.8 & 43.2 & 36.4 & 19.3 & 1.1 & 16.0 & 9.1 & 45.5 & 13.6 & 17.1 & 1.1 & 5.7 & 1.1 & 6.8 \\
\textit{S2Z} & 97.8 & 43.2 & 31.8 & 23.9 & 1.1 & 15.6 & - & 9.1 & 43.2 & 13.6 & 8.0 & 12.5 & 8.0 & 5.6 \\
\textit{N1Z} & 97.8 & 43.2 & 25.0 & 23.9 & 7.9 & - & 16.0 & 42.0 & 18.2 & 10.2 & 4.5 & 3.4 & - & 5.7 \\
\textit{N2Z} & 97.8 & 43.2 & 20.5 & 28.4 & 7.9 & - & 1.1 & 14.8 & 37.5 & 21.5 & 8.0 & 8.0 & 8.0 & 1.1 \\ \hline
\end{tabular}
\end{table*}

\begin{table*}[!t]
\renewcommand{\arraystretch}{1.3}
\caption{Use of \textit{Kitas} to Solve the Series of Table \ref{tabIQ} (\%).}
\label{tabIQres2}
\centering
\begin{tabular}{ccccccccccccccc}
\hline
\textit{kita}: & \textit{ANA} & \textit{BASIC} & \textit{DGD} & \textit{DGE} & \textit{DIV} & \textit{DOP} & \textit{EXP} & \textit{FOC} & \textit{LOG} & \textit{ML} & \textit{RED} & \textit{RSYM} & \textit{SOE} & \textit{SSYM} \\ \hline 
\textit{S1Z} & 2.3 & 100.0 & 0.0 & 0.0 & 17.0 & 2.3 & 1.1 & 26.1 & 2.3 & 10.2 & 17.0 & 0.0 & 0.0 & 0.0 \\
\textit{S2Z} & 2.3 & 100.0 & 0.0 & 0.0 & 17.0 & 2.3 & 1.1 & 23.9 & 2.3 & 10.2 & 21.6 & 1.1 & 0.0 & 1.1 \\
\textit{N1Z} & 0.0 & 100.0 & 0.0 & 0.0 & 11.4 & 3.4 & 5.7 & 26.1 & 1.1 & 20.5 & 27.3 & 0.0 & 0.0 & 1.1 \\
\textit{N2Z} & 0.0 & 100.0 & 0.0 & 0.0 & 10.2 & 3.4 & 8.0 & 15.9 & 2.3 & 20.5 & 30.7 & 8.0 & 0.0 & 2.3 \\ \hline 
\end{tabular}
\end{table*}

Table \ref{tabIQres1} shows the results of the four modes in which the algorithm was executed. The labels \textit{S1Z} and \textit{S2Z} correspond to the results using the BFS method stopping at the solution state, with constancy at one and two levels of the \textit{edk}, respectively. The labels \textit{N1Z} and \textit{N2Z} also correspond to the results using the BFS method, in this case without stopping at the solution state, with constancy at one level and two levels, respectively. In all of the execution modes, the solution to the problem was obtained in 97.8\% of the cases; that is, all of the series except for two of them. The solution found, which we will call type A solution, consists of an algorithmic pattern that is constituted by a list of \textit{kitas}, which allows us to reproduce all of the known terms of the series. Table \ref{tabIQres1} also shows the minimum number of elements needed to find the solution; that is, the other known elements are obtained by taking $n_{e}$ terms of the original series. In most series, this number varies between three and nine. This shows that, in general, the number of terms needed to find the underlying pattern is very small. The table also presents the depth of the graph used; that is, the percentage of cases in which it has been necessary to use from one to four \textit{kitas} to solve the problem. We can see that about half of the series need only one \textit{kita}. This indicates that, in general, they are relatively simple series. This depth is always greater when constancy is imposed on two levels of the \textit{edk} because more complex patterns result. Resolution times are always less than a second, with an average of 15 ms.

To solve this group of series, a set of 60 \textit{kitas} was used, which are formed by the types described in Section 3.2, applied with different parameters. Table \ref{tabIQres2} shows the percentage of use of each type of \textit{kita}, which provides a heuristic to improve the efficiency of the model. 

In conclusion, KitBit successfully solves almost all of this first set of non-trivial series based on IQ tests, using standard computer equipment in sub-second times. This result provides enough confidence in the model to deal with more complex problems, starting with those raised in previous models that appear in the literature.

\begin{table*}[!t]
\renewcommand{\arraystretch}{1.3}
\caption{Numerical Series Collected from the Literature.}
\label{tabotros}
\centering
\begin{tabular}{c c p{5cm} c c p{3.6cm} c c p{3.3cm}}
\hline
i & a & sequence & i & a & sequence & i & a & sequence\\ \hline
0 & \cite{AIS} & 5, 9, 35, 125, 345, 785, 1559, 2805, 4685 & 22 & \cite{RAG} & 5, 6, 7, 8, 10, 11, 14, 15 & 45 & \cite{BKC} & 2, 3, 5, 8, 13, 21, 34 \\
1 & \cite{AIS} & 2, 5, 11, 21, 37, 63, 107 & 23 & \cite{RAG} & 54, 48, 42, 36, 30, 24, 18 & 46 & \cite{BKC} & 1, 2, 1, 3, 1, 4, 1, 5 \\
2 & \cite{AIS} & 6, 288, 884736, 173946175488, & 24 & \cite{RAG} & 6, 8, 5, 7, 4, 6, 3, 5 & 47 & \cite{BKC} & 5, 10, 1, 2, 22, 44, 3, 6, 7, 14 \\
& & 2188749418902061056 & 25 & \cite{RAG} & 6, 9, 18, 21, 42, 45, 90, 93 & 48 & \cite{BKC} & 0, 1, 4, 9, 16 \\
3 & \cite{AIS} & 1, 2, 8, 48, 384, 3840 & 26 & \cite{RAG} & 7, 10, 9, 12, 11, 14, 13, 16 & 49 & \cite{BKC} & 1, 4, 9, 16, 25 \\
4 & \cite{AIS} & 5, 13, 35, 97, 275, 793, 2315, 6817 & 27 & \cite{RAG} & 8, 10, 14, 18, 26, 34, 50, 66 & 50 & \cite{BKC} & 4, 7, 12, 20, 32 \\
5 & \cite{AIS} & 12, 44, 144, 432, 1216, 3264, 8448, 21248 & 28 & \cite{RAG} & 8, 12, 10, 16, 12, 20, 14, 24 & 51 & \cite{BKC} & 4, 7, 12, 20, 33 \\
6 & \cite{MMI} & 1, 2, 3, 4, 5 & 29 & \cite{RAG} & 8, 12, 16, 20, 24, 28, 32, 36 & 52 & \cite{AMNS} & 7, 11, 15, 19, 23 \\
7 & \cite{MMI} & 1, 4, 9, 16 & 30 & \cite{RAG} & 9, 20, 6, 17, 3, 14, 0, 11 & 53 & \cite{AMNS} & 0, 2, 4, 6, 8, 10 \\
8 & \cite{MMI} & 1, 3, 9, 27 & 31 & \cite{HOF} & 0, 1, 4, 9 & 54 & \cite{AMNS} & 3, 6, 17, 66, 327 \\
9 & \cite{MMI} & 0, 1, 1, 2, 3, 5, 8, 13 & 32 & \cite{HOF} & 0, 2, 4, 6 & 55 & \cite{AMNS} & 1, 1, 2, 6, 24, 120, 720, 5040, \\
10 & \cite{MMI} & 1, 2, 4, 8, 16 & 33 & \cite{HOF} & 1, 1, 2, 3, 5 & & & 40320 \\
11 & \cite{RAG} & 12, 15, 8, 11, 4, 7, 0, 3 & 34 & \cite{HOF} & 0, 1, 2, 1, 4, 1 & 56 & \cite{AMNS} & 2, 5, 8, 11, 14, 17 \\
12 & \cite{RAG} & 148, 84, 52, 36, 28, 24, 22 & 35 & \cite{HOF} & 0, 0, 1, 1, 0, 0, 1, 1 & 57 & \cite{AMNS} & 3, 6, 12, 24, 48 \\
13 & \cite{RAG} & 2, 12, 21, 29, 36, 42, 47, 51 & 36 & \cite{HOF} & 0, 1, 3, 7 & 58 & \cite{AMNS} & 1, 2, 3, 5, 8, 13, 21, 34, 55 \\
14 & \cite{RAG} & 2, 3, 5, 9, 17, 33, 65, 129 & 37 & \cite{HOF} & 1, 2, 2, 3, 3, 3, 4, 4, 4, 4 & 59 & \cite{AMNS} & 1, 1, 2, 6, 24, 120 \\
15 & \cite{RAG} & 2, 5, 8, 11, 14, 17, 20, 23 & 38 & \cite{SCH} & 1, 4, 7, 10, 13, 16, 19, 22 & 60 & \cite{AMNS} & 1, 2, 3, 4 \\ 
16 & \cite{RAG} & 2, 5, 9, 19, 37, 75, 149, 299 & 39 & \cite{SCH} & 2, 4, 3, 5, 4, 6, 5, 7 & 61 & \cite{AMNS} & 3, 2, 1, 0 \\
17 & \cite{RAG} & 25, 22, 19, 16, 13, 10, 7, 4 & 40 & \cite{SCH} & 4, 11, 15, 26, 41, 67, 108, 175 & 62 & \cite{AMNS} & 1, 11, 111, 1111 \\
18 & \cite{RAG} & 28, 33, 31, 36, 34, 39, 37 & 41 & \cite{SCH} & 5, 6, 12, 19, 32, 52, 85, 138 & 63 & \cite{AMNS} & 1, 44, 1, 27, 1, 92 \\
19 & \cite{RAG} & 3, 6, 12, 24, 48, 96, 192 & 42 & \cite{SCH} & 8, 10, 14, 18, 26, 34, 50, 66 & 64 & \cite{AMNS} & 1, 39, 1, 35, 1, 28 \\
20 & \cite{RAG} & 3, 7, 15, 31, 63, 127, 255 & 43 & \cite{SCH} & 1, 2, 2, 3, 3, 3, 4, 4, 4, 4, 5 & 65 & \cite{AMNS} & 1, 1, 7, 1 \\
21 & \cite{RAG} & 4, 11, 15, 26, 41, 67, 108 & 44 & \cite{BKC} & 2, 2, 4, 8, 32, 256 & 66 & \cite{AMNS} & 46, 147, 9, 1, 1, 1 \\ \hline
\end{tabular}
\end{table*}

\begin{table*}[!t]
\renewcommand{\arraystretch}{1.3}
\caption{Results Obtained for the Series in Table \ref{tabotros}.}
\label{tabotrosres1}
\centering
\begin{tabular}{ccccccccccccccccc}
\hline
& \multicolumn{2}{c}{results (\%)} & \multicolumn{4}{c}{depth of the graph} & time (ms) & \multicolumn{8}{c}{$n_e$ (\%)} \\ \cline{2-16}
& solved & solved+ & 1 (\%) & 2 (\%) & 3 (\%) & 4 (\%) & $t_{med}$ & 3 & 4 & 5 & 6 & 7 & 8 & 9 & 10 \\ \hline 
\textit{S1Z} & 88.1 & 91.0 & 26.2 & 47.6 & 21.3 & 4.9 & 30.6 & 32.8 & 29.5 & 8.2 & 18.0 & 6.6 & 1.6 & - & 3.3 \\
\textit{S2Z} & 73.1 & 91.0 & 21.3 & 32.8 & 36.1 & 9.8 & 34.0 & 8.2 & 29.5 & 19.7 & 14.8 & 19.7 & 3.3 & 1.6 & 3.2 \\
\textit{N1Z} & 89.6 & 91.0 & 26.2 & 32.8 & 34.4 & 6.6 & - & 41.0 & 23.0 & 23.0 & 4.9 & 4.9 & 3.2 & - & - \\
\textit{N2Z} & 83.6 & 91.0 & 21.3 & 26.2 & 31.2 & 21.3 & - & 13.1 & 31.1 & 16.4 & 21.3 & 11.5 & 6.6 & - & - \\ \hline
\end{tabular}
\end{table*}

\begin{table*}[!t]
\renewcommand{\arraystretch}{1.3}
\caption{Use of \textit{Kitas} to Solve the Series of Table \ref{tabotros} (\%).}
\label{tabotrosres2}
\centering
\begin{tabular}{ccccccccccccccc}
\hline
\textit{kita}: & \textit{ANA} & \textit{BASIC} & \textit{DGD} & \textit{DGE} & \textit{DIV} & \textit{DOP} & \textit{EXP} & \textit{FOC} & \textit{LOG} & \textit{ML} & \textit{RED} & \textit{RSYM} & \textit{SOE} & \textit{SSYM} \\ \hline 
\textit{S1Z} & 0.0 & 100.0 & 0.0 & 0.0 & 29.5 & 1.6 & 3.3 & 24.6 & 0.0 & 21.3 & 21.3 & 1.6 & 1.6 & 0.0  \\
\textit{S2Z} & 0.0 & 100.0 & 0.0 & 3.3 & 26.2 & 1.6 & 6.6 & 19.7 & 0.0 & 32.8 & 39.3 & 3.3 & 1.6 & 0.0 \\
\textit{N1Z} & 0.0 & 100.0 & 0.0 & 0.0 & 14.8 & 4.9 & 4.9 & 24.6 & 0.0 & 36.1 & 32.8 & 1.6 & 1.6 & 0.0 \\
\textit{N2Z} & 0.0 & 100.0 & 0.0 & 0.0 & 23.0 & 8.2 & 8.2 & 11.5 & 0.0 & 42.6 & 44.3 & 13.1 & 1.6 & 0.0 \\ \hline 
\end{tabular}
\end{table*}

%%%%%%%%%%%%%%%%%%%%%%%%%%%%%%%%%%%%%%%%%%%%%%%%%%%%%%%%%%%%%%%%%%%%%%%%%%%%%%%%%%%%%%%%%%%%%%%%%%%%%%%%%%%%%
\subsection{Series from previous models} \label{s5_2}

KitBit was then tested with a new set of 67 series compiled from different articles dedicated to proposing procedures for solving numerical series \cite{AIS}, \cite{MMI}, \cite{RAG}, \cite{HOF}, \cite{SCH}, \cite{BKC} and \cite{AMNS}. The list of the problems is summarized in Table \ref{tabotros}. Column \textit{i} indicates the index in this dataset and \textit{a}  indicates the reference article. KitBit approaches this list using the same set of \textit{kitas} as in the previous subsection. The results are shown in Tables \ref{tabotrosres1} and \ref{tabotrosres2}.

Table \ref{tabotrosres1} shows the results for this set of number sequences in the four execution modes that were given earlier. The number of solved problems is between 73.1\% and 89.6\%, and is able to improve up to 91\% in all cases if we increase the number of terms of some series. In other words, the model solves the entire set of series evaluated with a higher success rate than previous models, except for six of them. In this case, the solutions are also of type A; that is, in all cases an algorithmic pattern is found that includes all the terms of the series. The depth of the graph shifts towards the use of a greater number of \textit{kitas}, which indicates the greater complexity of the series when compared with the previous set. The number of elements necessary to solve the problem is shown in Table \ref{tabotrosres1}. Between three and 10 elements are necessary to find the solution, as we found with the previous group. Table \ref{tabotrosres2} shows the percentage of use of each type of \textit{kita}. This shows that some that had not been used with the previous group of series are relevant in this case. Resolution times were also similar to the previous group, always below one second, with an average time of 34 ms.

We can make several observations if we compare these results with those presented in the publications from which the sequences were extracted. Our system obtains the pattern for the series resolved in \cite{AIS}. KitBit is able to obtain the solution of all the series evaluated in \cite{MMI}, while the authors fail to obtain some of them using two different methodologies. KitBit manages to correctly solve  the 20 problems posed in \cite{RAG}, while its authors solve 17 of them and the IGOR2 model \cite{HOF} manages to solve 14. Likewise, KitBit algorithms also solve all of the series proposed to test the IGOR2 model \cite{SCH}, even the one that this model fails to obtain. For the series proposed to test SeqSolver \cite{BKC}, KitBit can solve all of them, except one. Finally, regarding the series proposed to validate the ASolver \cite{AMNS} method, KitBit finds a pattern for all but five of them. In conclusion, KitBit is capable of solving the numerical series of IQ tests reported in the bibliography, as well as other series not related to the IQ tests, and performs better than all of the methods that have previously been proposed.

%%%%%%%%%%%%%%%%%%%%%%%%%%%%%%%%%%%%%%%%%%%%%%%%%%%%%%%%%%%%%%%%%%%%%%%%%%%%%%%%%%%%%%%%%%%%%%%%%%%%%%%%%%%%%

\subsection{Series from OEIS database} \label{s5_3}

OEIS is the largest database of the entire series, which compiles all available information about them and is widely cited in the literature. Its size is continuously growing and, at the time of our tests, it contained a total of 347,736 sequences. The repeated sequences were eliminated from this initial set, as well as those that have less than three elements or less than four if they are formed by the repetition of the same element. The sample on which the KitBit algorithms were finally run was 341,553 entire series. The BFS technique was used without stopping at the solution state, with single-level constancy of the \textit{edks} and an epsilon of $\epsilon = e^{-18}$. Initially, 55 \textit{kitas} and a state tree depth equal to two were used. Next, a depth equal to three was used on unsolved problems. Again, on the unsolved cases, a depth equal to four was applied. In this case we reduced the number of \textit{kitas} to 34 and 32 to limit the total resolution time of this large set of problems. As in the previous data sets, the resolution of each sequence was carried out using the minimum number of elements necessary to find the pattern capable of reproducing the complete sequence as it appears in the OEIS database. Given the large number of series to be solved, the algorithms were run using Google's free cloud resources through Google Colab.

We start by addressing the resolution of the 341,553 series, finding the solution in 87,514 cases, or 25.6\% of the total; as shown in Table \ref{taboeis1}. This is the largest number of series resolved to date for the OEIS database. Previously, 26,951 sequences (7.89\% of the database) were reported from a selection of 57,524 series with at least 20 elements and values less than 1,000 \cite{RAG}. The authors of this study considered a series solved when they are `able to correctly predict the last number of the series'. In our case, we are always able to predict a larger number of elements, even the entire series, depending on the \textit{kitas} used to obtain the solution. In this way, we find two types of solutions: type A solutions (as mentioned previously), and a new type of solutions that we will call type B. Type A solutions are those in which an algorithmic pattern is obtained that allows us to reproduce the sequence starting from a given term, using all previous terms up to the first to get that pattern. For this type, KitBit has managed to solve 28,293 series, or 8.3\% of the total. Type B solutions are those in which we are capable of reproducing the sequence starting from a certain term. However, the pattern obtained has not taken into account all of the previous terms but rather a finite number of them. To this last type fundamentally belong the solutions that use \textit{SSYM} that, by not using all the initial elements of the series, allow forward prediction but do not include these in the pattern found. KitBit has managed to find a type B solution for 59,221 cases, or 17.3\% of the total, where the number of elements $n$ that it is able to predict depends on the complexity of the series and the number of known elements. In type B solutions, the next term is predicted in 44.4\% of the cases and from two to nine terms in 45.7\% of them. In 9.9\% of the remaining solutions, more than 10 terms are obtained. For type A solutions, more than 10 terms are predicted in 65\% of them. The extrapolation of the series beyond the known elements, for both type A and type B solutions, depends on the \textit{kitas} used and the pattern found.

By analyzing the unresolved series, we find that a high percentage of them fall into the following categories: a) series based on complex mathematical functions that include trigonometric, hyperbolic and logarithmic functions; b) series that grow very quickly, such as exponential functions or $n^{th}$ powers, where the value between contiguous elements of the series increases by one order of magnitude; c) series based on prime numbers; and d) series based on combinations of other series. Among these categories, we find that 4.6\% of series contain very large and very small numbers. Although these are very difficult for our algorithms to solve, they are capable of solving some of them (0.3\%). The series that result from the combination of others are 17.4\% of the total and suppose an artificial degree of difficulty that, at this time, our system does not intend to address. Series based on prime numbers represent 16\%, which require new algorithms to be incorporated. Even with this degree of difficulty, KitBit is able to solve 51,168 out of 240,617 series of this type, or 21.3\% of them. Out of these categories, the OEIS database contains 100,936 series, for which KitBit finds solutions in 36\% of them; as shown in Table \ref{taboeis1}. By further restricting the selection criteria, as in other models, the results would improve. Likewise, it is expected that a greater number of series would be resolved by expanding the set of \textit{kitas} and the depth of the graph.

\begin{table*}[!t]
\renewcommand{\arraystretch}{1.3}
\caption{Results for the Series in the OEIS Database.}
\label{taboeis1}
\centering
\begin{tabular}{cccccc}
\hline
& Total series & Type A & Type B & Type A + Type B \\ \hline 
Entire database & 341553 & 28293 (8.3\%) & 59221 (17.3\%) & 87514 (25.6\%) \\ 
Selection & 100936 & 11938 (11.8\%) & 24408 (24.2\%) & 36346 (36.0\%) \\ \hline
\end{tabular}
\end{table*}

\begin{table*}[!t]
\renewcommand{\arraystretch}{1.3}
\caption{Most Frequent Patterns in the Type A Series. $n_s$ is the Number of Series Solved Using this Pattern.}
\label{taboeis2}
\centering
\begin{tabular}{cccccc}
\hline
pattern & $n_s$ & pattern & $n_s$ & Pattern & $n_s$ \\ \hline 
\textit{BASIC} (only) & 6623 & \textit{ML}(2,1), \textit{ML}(0,1), \textit{FOC}(0,\{1,1\}) & 310 & \textit{RED}(6), \textit{DGD} & 232 \\
\textit{FOC}(0,\{1,1\}) & 1992 & \textit{RED}(1), \textit{DIV} & 309 & \textit{RED}(4), \textit{DGD} & 213 \\
\textit{ML}(1,1), \textit{ML}(0,1), \textit{FOC}(0,\{1,1\}) & 1289 & \textit{ANA}(0,5) & 303 & \textit{RED}(7), \textit{DGD} & 212 \\
\textit{FOC}(0,\{1,1,1,1\}) & 873 & \textit{ANA}(0,3) & 285 & \textit{FOC}(0,\{1,1,1\}), \textit{RED}(1), \textit{DIV} & 206 \\
\textit{FOC}(0,\{1,1,1\}) & 628 & \textit{RED}(1), \textit{DGE}, \textit{FOC}(0,\{1,1\}) & 278 & \textit{RSYM} & 195 \\
\textit{RED}(1), \textit{RSYM} & 564 & \textit{RED}(1), \textit{ML}(0,1) & 275 & \textit{FOC}(0,\{1,1\}), \textit{RED}(1), \textit{DIV} & 179 \\
\textit{ML}(0,1), \textit{FOC}(0,\{1,1\}) & 488 & \textit{ML}(1,1) & 272 & \textit{RED}(2), \textit{DOP}($\times$,$+$) & 178 \\
\textit{RED}(1), \textit{ML}(0,1), \textit{FOC}(0,\{1,1\}) & 429 & \textit{ANA}(0,4) & 269 & \textit{FOC}(0,\{1,1,1\}), \textit{FOC}(0,\{1,1,1,1\}) & 177 \\
\textit{ANA}(0,2) & 370 & \textit{FOC}(0,\{1,1,1\}), \textit{FOC}(0,\{1,1\}) & 236 & others & 10908\\
\hline
\end{tabular}
\end{table*}

\begin{table*}[!t]
\renewcommand{\arraystretch}{1.3}
\caption{Examples of Solved Sequences.}
\label{taboeis3}
\centering
\begin{tabular}{cccc}
\hline
OEIS index & sequence & pattern & prediction \\ \hline
A071420 & 7, 8, 5, 5, 3, 4, 4, 6, 9, 7, 8, 8, 7, 8, 5, 5, 3, 4, 4, 6, 9, 7, 8, 8 & \textit{FOC}(0,\{1,1,1\}), \textit{FOC}(0,\{1,1,1,1\}) & 7, 8, 5, 5, 3, 4, 4, 6, 9 \\
\multirow{2}{*}{A082310} & \multirow{2}{*}{0, 1, 7, 57, 455, 3641, 29127, 233017} & \multirow{2}{*}{\textit{FOC}(0,\{1,1\}), \textit{RED}(1), \textit{DIV}} & 1864135, 14913081, 119304647, \\
 & & & 954437177 \\
A121294 & 1, 2, 5, 8, 10, 12, 17, 22, 27, 30, 33, 36, 43, 50, 57, 64, 68 & \textit{RED}(1), \textit{DGE}, \textit{FOC}(0,\{1,1\}) & 72, 76, 80, 89, 98, 107, 116, 125 \\ 
\multirow{2}{*}{A194385} & 6, 12, 18, 24, 30, 36, 228, 234, 240, 246, 252, 258, & \multirow{2}{*}{\textit{RED}(6), \textit{DGD}} & \multirow{2}{*}{690, 696, 702, 708, 714, 720} \\
 & 264, 456, 462, 468, 474, 480, 486, 492, 684 & & \\
A227589 & 1, 4, 7, 12, 16, 23, 29, 38, 46, 57, 67, 80, 92, 107, 121 & \textit{ML}(1,1), \textit{ML}(0,1), \textit{FOC}(0,\{1,1\}) & 138, 154, 173, 191, 212, 232 \\ \hline
\end{tabular}
\end{table*}

The variety of series collected in OEIS, many of which are highly complex, is reflected in the depth of the state tree. We observe that for depths of 1, 2, 3 and 4 levels, the percentage of resolved sequences is 23.4\%, 21.4\%, 28.2\% and 27.0\%, respectively. As we can see, a quarter of the resulting sequences have required up to a fourth level of depth, this being the maximum allowed in the execution of the model. In addition, the most used combinations of \textit{kitas} are shown in Table \ref{taboeis2}. The most used are \textit{FOC} (56.9\%), \textit{ML} (32.7\%), \textit{RED} (28.7\%) and \textit {ANA} (9 .8\%), plus \textit{BAS}. This result contrasts with those obtained for the two groups of the series of IQ tests, due to the greater complexity of the series in OEIS.

Finally, Table \ref{taboeis3} shows several examples showing the reference in OEIS, the terms used to find the underlying pattern, the pattern itself as a list of \textit{kitas}, and the prediction. These examples, like many of the series within OEIS, are non-trivial and serve to demonstrate the model's ability to find the underlying pattern and predict the next terms in sequences with this degree of difficulty. Consequently, KitBit is a computational system that is capable of solving the largest number of sequences contained in OEIS to date, well above the maximum value reported in the literature.

%%%%%%%%%%%%%%%%%%%%%%%%%%%%%%%%%%%%%%%%%%%%%%%%%%%%%%%%%%%%%%%%%%%%%%%%%%%%%%%%%%%%%%%%%%%%%%%%%%%%%%%%%%%%%%%%%%%%%%%%%%%%%%%%%%%%%%%%%%%%%%%%%%%%%%%%%%%%%%%%%%%%%%%%%%%%%%%%%%%%%%%%%%%%%%%%%%%%%%%%%%%%%%%%%%%%%%%%%%%%

\section{Conclusions} 

We have presented the fundamentals of KitBit, which is a new computational model capable of obtaining the underlying pattern in series of numbers, as well as its efficiency in solving IQ tests and a series of greater complexity, such as those found in the OEIS database. KitBit uses a set of algorithms or \textit{kitas} that perform different operations on the numerical series, most of them non-analytical, to find the underlying pattern without the need for training or prior knowledge, in very short times, always less than a second, using standard computing power. This pattern is represented by a sequence of \textit{kitas} instead of a mathematical function, which allows us to reproduce the known elements of the numerical sequence and extrapolate to new ones.

KitBit has managed to solve 97.8\% of a first set of 90 series associated with IQ tests collected from different sources. In addition, it solves 91\% of a second set of 67 series compiled from previous publications aimed at solving numerical series and intelligence tests, surpassing the results of these methods. Finally, KitBit deals with the resolution of 341,553 entire series included in OEIS database and is the only model to date that considers the resolution of its entirety. KitBit is capable of solving 87,514 sequences, or 25.6\% of the total. This is the largest number of solved series to date. Within this result we find two types of solutions: a first group in which the sequence is specified from a certain term, using the pattern found in all the previous terms up to the first; and a second group, in which the sequence is also specified from a certain term but the pattern found does not include all the previous terms up to the first.

These results are promising to face other types of problems using the same methodology, such as the resolution of other types of IQ test or more specific complex problems through the development of new \textit{kitas}, which will be presented in future publications.
\vspace{20px}
%%%%%%%%%%%%%%%%%%%%%%%%%%%%%%%%%%%%%%%%%%%%%%%%%%%%%%%%%%%%%%%%%%%%%%%%%%%%%%%%%%%%%%%%%%%%%%%%%%%%%%%%%%%%%%%%%%%%%%%%%%%%%%%%%%%%%%%%%%%%%%%%%%%%%%%%%%%%%%%%%%%%%%%%%%%%%%%%%%%%%%%%%%%%%%%%%%%%%%%%%%%%%%%%%%%%%%%%%%%%
\ifCLASSOPTIONcaptionsoff
  \newpage
\fi
\bibliographystyle{IEEEtran}
\bibliography{IEEEabrv,biblio_edited.bib}
\vspace*{-6\baselineskip}
\begin{IEEEbiography}[{\includegraphics[width=25mm,height=32mm]{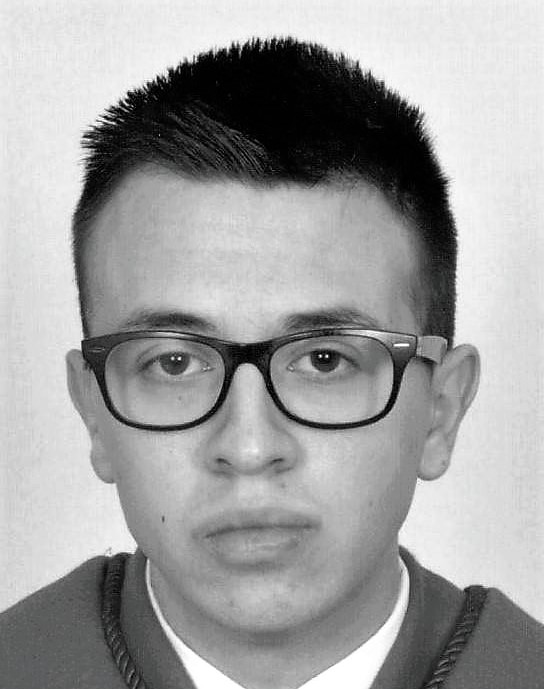}}]{Víctor Corsino}
is a PhD student in Science and Technology Applied to Industrial Engineering, Universidad de Castilla La-Mancha (UCLM), Spain. He has a degree in electronic engineering from this university, with a mention in advanced electronic technologies, and a Master's degree in artificial intelligence from Universidad Politécnica de Madrid
(UPM). His current research interests include pattern recognition, machine learning and microsensors.
\end{IEEEbiography}
\vspace*{-7\baselineskip}

\begin{IEEEbiography}[{\includegraphics[width=25mm,height=32mm]{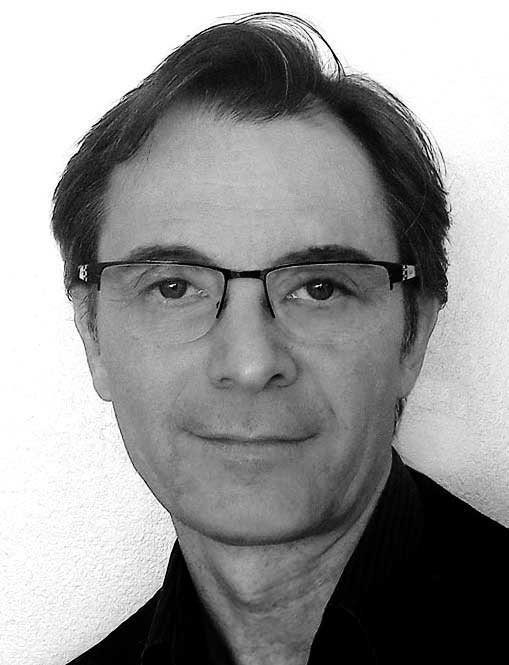}}]{José Manuel Gilpérez}
is Full Time Professor at the School of Industrial and Aerospace Engineering (EIIA) in Toledo, Universidad de Castilla La-Mancha (UCLM), Spain. His current research interests are focused on applied artificial intelligence in different fields, such pattern recognition, machine learning, neurobiology and genetics in partnership with several universities and research centers. 
\end{IEEEbiography}

\vspace*{-6\baselineskip}
\begin{IEEEbiography}[{\includegraphics[width=25mm,height=32mm]{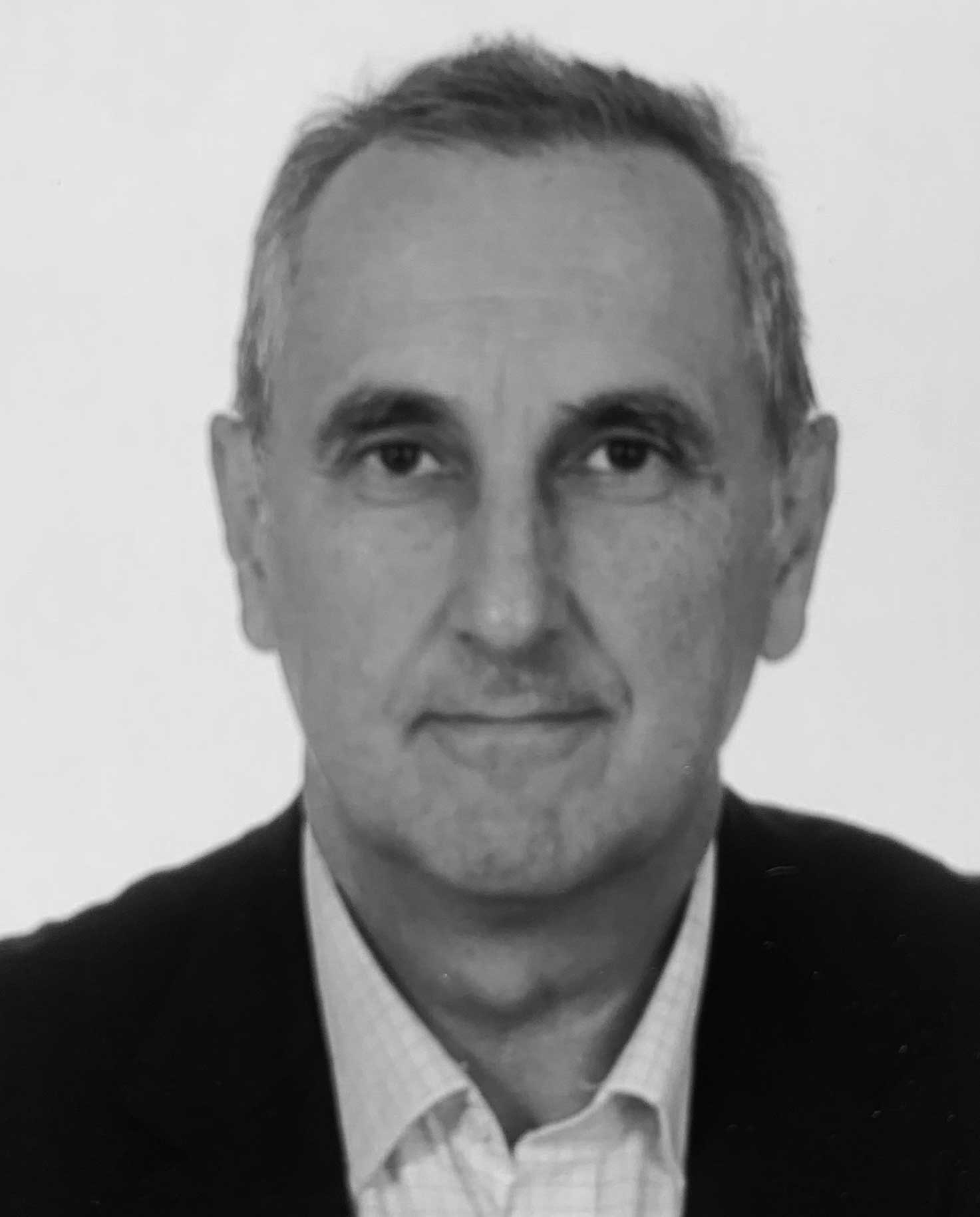}}]{Luis Herrera}
has a BS in Aerospace Engineering from Universidad Politécnica de Madrid (UPM). He holds a Certificate in Exponential Technologies from Singularity University. He is CEO of Smart Technologies Investments, which is the company that is developing the KitBit project.
\end{IEEEbiography}

% You can push biographies down or up by placing a \vfill before or after them. 
\end{document}